\documentclass{WileyMSP-template}
\usepackage{multirow}
\usepackage{algorithmic}
\usepackage{algorithm}
\usepackage{array}
\usepackage[caption=false,font=normalsize,labelfont=sf,textfont=sf]{subfig}
\usepackage{textcomp}
\usepackage{stfloats}
\usepackage{url}
\usepackage{verbatim}
\usepackage{booktabs}
\usepackage{graphicx}
\usepackage{subfig}
\usepackage{cite}
\usepackage{url}
\usepackage{color}
\usepackage{amsmath}
\usepackage{afterpage}
\usepackage{rotating}
\usepackage{lscape}
\usepackage[dvipsnames]{xcolor}
\usepackage{xr}
\usepackage{hyperref}

\begin{document}

\pagestyle{fancy}
\rhead{\includegraphics[width=2.5cm]{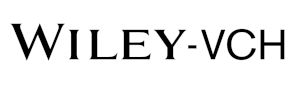}}

\title{Navigating Intelligence: A Survey of Google OR-Tools and Machine Learning for Global Path Planning in Autonomous Vehicles}

\maketitle


\author{Alexandre Benoit}
\author{Pedram Asef}

\begin{affiliations}
Alexandre Benoit \\
University of Bath \\
amgb20@bath.ac.uk \\
ORCID: 0009-0002-2886-3513

Pedram Asef \\
University College London \\
pedram.asef@ucl.ac.uk \\
ORCID: 0000-0003-3264-7303

\end{affiliations}


\keywords{Autonomous Vehicles, Global Path Planning, Google OR-Tools, Machine Learning, Q-learning Algorithm, Reinforcement Learning, Travelling Salesman Problem.}

\begin{abstract}

This study offers an in-depth examination of Global Path Planning (GPP) for unmanned ground vehicles (UGV), focusing on an autonomous mining sampling robot named ROMIE, which plays a crucial role in geochemical mining sampling. GPP is essential for ROMIE's optimal performance, as it involves solving the Traveling Salesman Problem (TSP), a complex graph theory challenge that is crucial for determining the most effective route to cover all sampling locations in a mining field. This problem is central to enhancing ROMIE's operational efficiency and competitiveness against human labor by optimizing cost and time.
The primary aim of this research is to advance GPP by developing, evaluating, and improving a cost-efficient software solution and web application. We delve into an extensive comparison and analysis of various Google OR-Tools optimization algorithms, designed to address different TSP scenarios. Our study is driven by the goal to not only apply but also test the limits of OR-Tools' capabilities by integrating fundamental Reinforcement Learning techniques like Q-Learning and Double Q-Learning into our approach. This enables us to compare these basic methods with OR-Tools, assessing their computational effectiveness and real-world application efficiency.
Our comparative analysis seeks to provide insights into the effectiveness and practical application of each technique, informing future advancements in GPP software. Our findings indicate that Q-Learning stands out as the optimal strategy, demonstrating superior efficiency by deviating only 1.2\% on average from the optimal solutions across our datasets. In conclusion, the research shows that Q-Learning-based algorithms are the most effective, suggesting their significant potential in delivering cost-efficient and robust solutions in real-world mining operations, thereby enhancing the capabilities of autonomous robot like ROMIE.
 \break
\end{abstract}


\section{Introduction}

\begin{figure}[H]
    \centering
    \includegraphics[width=\columnwidth, height=0.4\columnwidth, keepaspectratio]{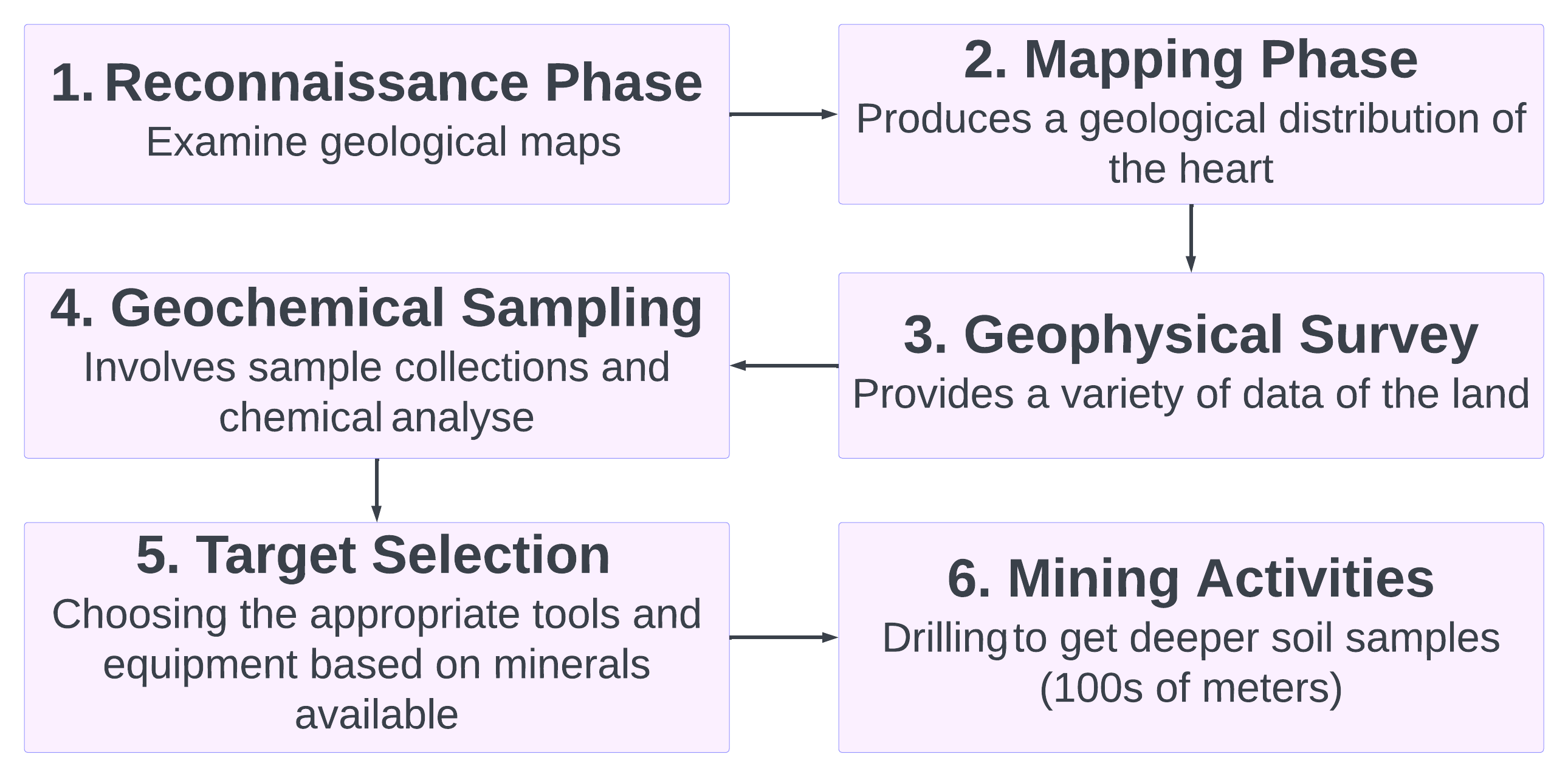}
    \caption{Diagram of hierarchical sampling processes before undergoing mining activities.}
    \label{fig:sampling}
\end{figure}

ROMIE (Robotic Ore Mineral Identification and Exploration) is a pioneering robot specializing in tackling the field of exploration and geochemical sampling within the mining industry. Undertaken before mining activities (Fig.~\ref{fig:sampling}), the geochemical sampling process entails the collection and analysis of soil samples to identify potential mineral deposits. Traditionally relying on manual labor, the existing methods of sampling pose significant issues in terms of time, cost, working conditions, and sustainability. The process is labor-intensive and demands a substantial workforce, placing a burden on related companies. \break

\begin{figure}[H]
\centering
\includegraphics[width=\columnwidth, height=0.45\columnwidth, keepaspectratio]{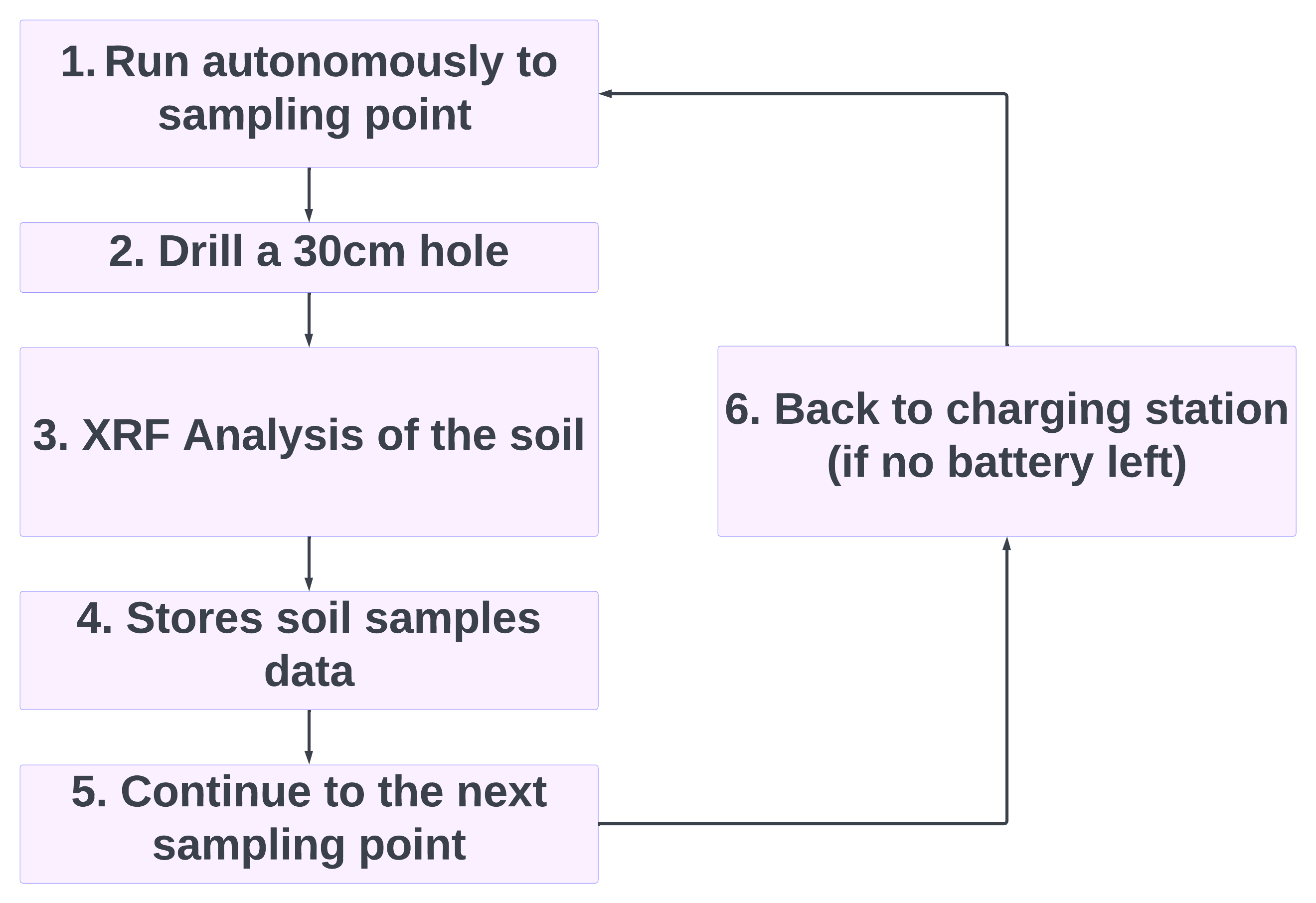}
\caption{ROMIE’s functioning procedure.}
\label{fig:Functioning}
\end{figure}

\begin{figure}[H]
\centering
\includegraphics[width=\columnwidth, height=0.65\columnwidth, keepaspectratio]{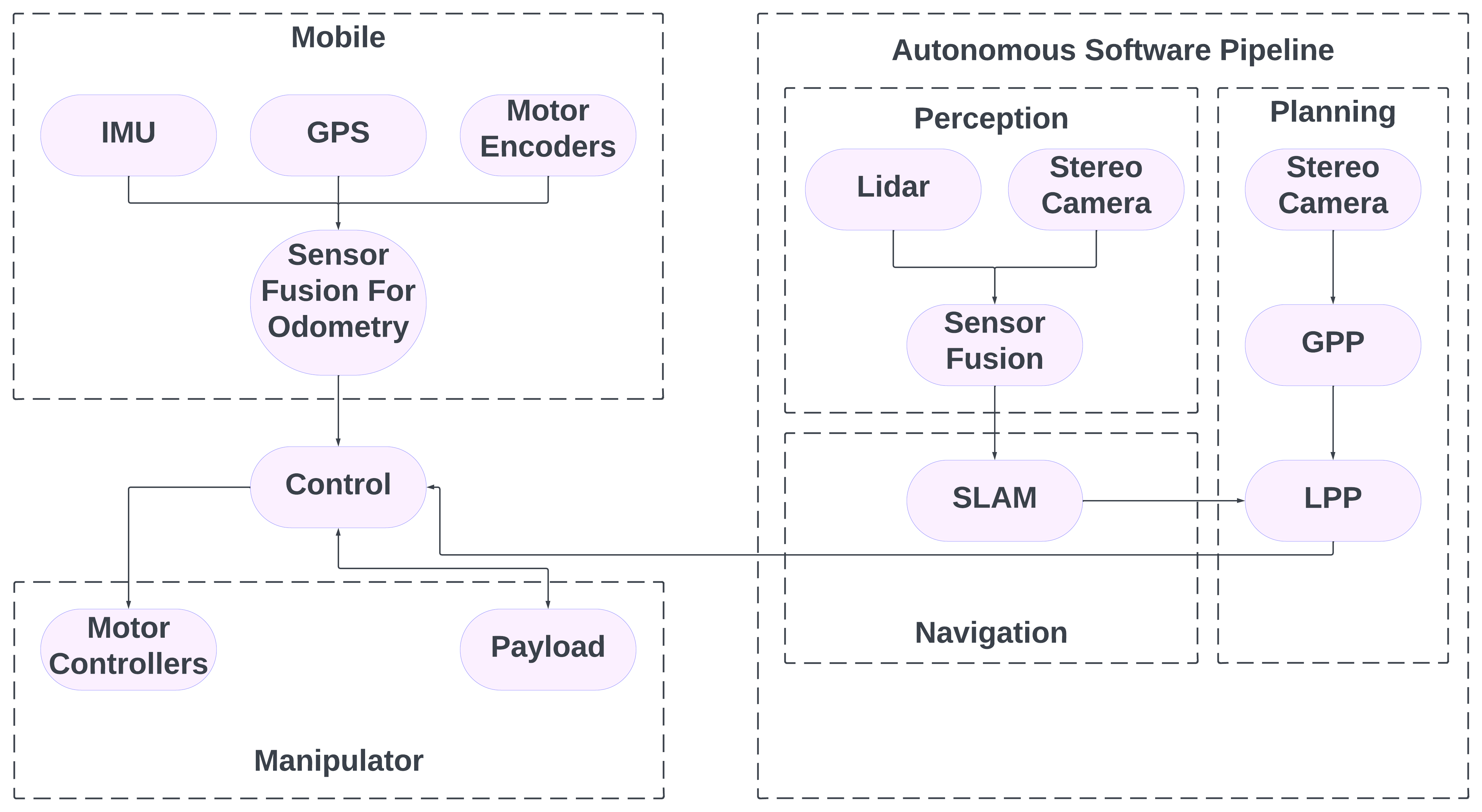}
\caption{Overall structure for ROMIE; LPP: Local Path Planning.}
\label{fig:Strucutre}
\end{figure}

To date, the process of sampling kilometers of fields in the mining industry has been performed manually becoming a laborious task. In response to this challenge, the ROMIE software and hardware have been investigated and developed as an innovative autonomous global path planning system to optimize the navigation of mobile robots in a variety of environments. The system utilizes advanced algorithms and state-of-the-art hardware to generate efficient and reliable paths, taking into account obstacles, constraints, and objectives (Fig.~\ref{fig:Strucutre}). This is performed by a state-of-the-art object detection algorithm supported by two stereo cameras as well as a detailed mapping executed by a LIDAR. ROMIE traverses the field to go to specific point having mineral of interest, drill and analyse the soil with an on-board earth auger and X-ray fluorescence (XRF) (Fig.~\ref{fig:Functioning}). \break

Solving the Global Path Planning (GPP) problem, is analogous to solving the Travelling Salesman Problem (TSP) \cite{uavpp}. The TSP is a well known NP-hard problem \cite{wikitsptsp} in combinatorial problems, important in theoretical computer science \cite{CS} and optional research \cite{OR}. It has become common practise to use optimization algorithms which are broadly classified as deterministic, yielding consistent results for the same inputs, and stochastic, which use randomness to explore multiple solution paths \cite{deter}. Stochastic algorithms, subdivided into heuristic and meta-heuristic types, are particularly useful for challenging optimization problems due to their exploratory methods and practicality. The TSP is a prime example where heuristic approaches often deliver near-optimal solutions efficiently, making them suitable for NP-hard problems. The choice of algorithm, however, depends significantly on the application's specifics. \break

Numerous renowned methods, e.g., Genetic Algorithm \cite{GAalgo}, Tabu Search \cite{TabuSearch}, Particle Swarm Optimization \cite{PCO}, Simulated Annealing \cite{SAaaa}, and Ant Colony Optimization \cite{ACO}, have been applied to approximate the optimal solution to the TSP. \break

More recently, machine learning (ML) methods have emerged as viable alternatives, offering solutions that are both computationally efficient and closer to the exact solution. Specifically, reinforcement learning (RL) has demonstrated its utility as a straightforward and accessible technique to address the TSP \cite{accesible}. RL is a branch of ML where an agent maximizes rewards through trial-and-error interactions with its environment. Essential this is the Markov Decision Process, where decisions rely on the present state, embodying all relevant past information, thus simplifying the learning process \cite{RLdef}. \break

The contribution and novelty of this paper can be stated as follow: the GPP framework is a new software of its kind, for autonomous sampling robots with a unique comparative analysis of real-time optimization algorithms, namely, Google OR-Tools and RL algorithms, such as Q-Learning and Double Q-Learning. The main objective is to identify the optimal TSP tour among the sampling points in real-time operations. The software prioritizes the minimization of travel distance over computational time efficiency because a shorter distance translates into less prospecting time, resulting in huge cost savings. Finally, the developed ROMIE software is physically developed and experimentally validated. \break

The interaction between GPP and Local Path Planning (LPP) is a critical component in our robotic navigation system. The GPP is tasked with determining the most efficient route for the robot to traverse through a predefined set of waypoints. This process involves calculating the shortest path that the robot should take to visit each designated sampling point, based on a discrete set of coordinates.\break

\begin{figure}[H]
\centering
\includegraphics[width=\columnwidth, height=0.75\columnwidth, keepaspectratio]{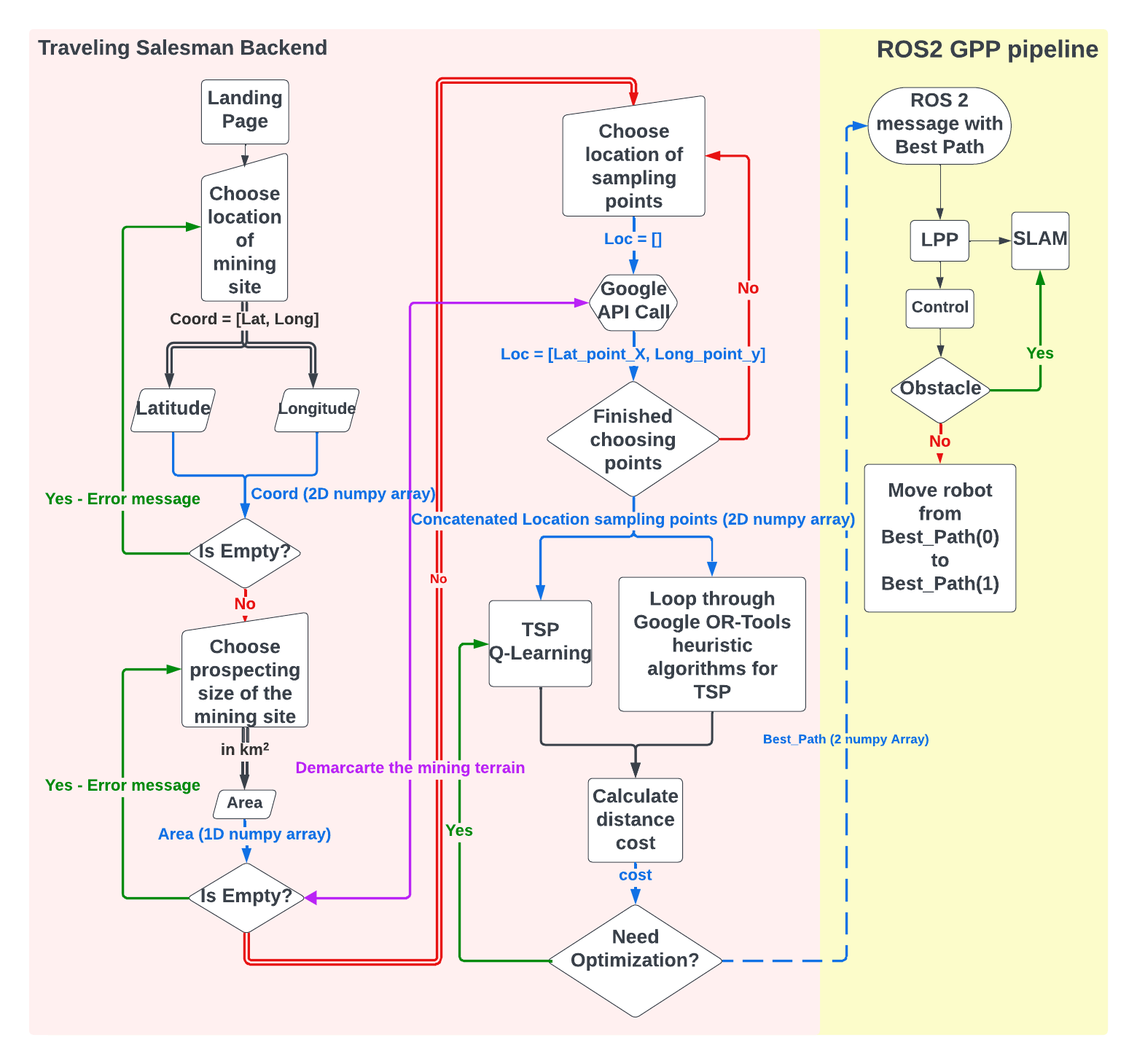}
\caption{Overall interaction from GPP to LPP}
\label{fig:LPPGPP}
\end{figure}

Once the GPP has computed this path, it communicates the sequence of waypoints to the LPP system using a ROS2 message that contains the list of coordinates (Fig.~\ref{fig:LPPGPP}). The LPP's role is then to guide the robot from one waypoint to the next, handling real-time navigation challenges such as obstacle avoidance. It's important to note that the LPP is equipped with sensors and cameras to detect and navigate around obstacles, adding necessary detours to the planned route. These adjustments are made independently of the GPP, which remains focused on the overarching route between waypoints. \break

As mentioned in Fig.~\ref{fig:LPPGPP}, the GPP receives 2D coordinates and the consideration of terrain elevation has been neglected for the context of our application—mining operations predominantly in regions like Africa and Australia where we assume operation on relatively flat terrain. This assumption is based on the typical topography of these areas and the economic feasibility of mining operations on level ground. Therefore, while elevation could impact the path planning in three-dimensional space, our current framework does not incorporate elevation data, focusing instead on optimizing paths on a two-dimensional plane. \break

This approach allows us to streamline the GPP's functionality, concentrating on the primary goal of efficient waypoint navigation, while the LPP focuses on the intricacies of immediate, local movements and obstacle avoidance.\break

We aim to achieve a comparative and review study of Google OR-Tools and simple Reinforcement Learning (Q and double Q-Learning) applied to the TSP. It was achieved by demonstrating in this paper the following several objectives:
\begin{enumerate}
    \item \textbf{Efficient Route Identification for GPP}: Our first objective is to determine the most cost-effective TSP route for ROMIE, focusing on minimizing travel distance over computational time efficiency. This approach is motivated by the fact that shorter distances equate to less prospecting time, leading to substantial cost savings.For the same size of land, ROMIE is \textbf{6} times faster and \textbf{1.5} cheaper than human labour. Finally, the developed ROMIE software is physically developed and experimentally validated.
    
    \item \textbf{Development of Affordable GPP Software and Experimenting Google OR-Tools Web Interface Playground}: We are developing an interactive Google Map interface, using their API, to enable mining geologists and engineers to manually select specific sampling points. This tool will also serve as an experimental platform for evaluating Google OR-Tools (GOT) and basic ML techniques in a TSP context for educational/academia purposes. Our software integrates effectively with Google Maps and OR-Tools, providing an end-to-end solution for our web app's visual interface.

    \item \textbf{Employment of Various Optimization Techniques }: We are utilizing Google OR-Tools API for optimization and comparing its performance with foundational ML strategies, exploring their practicality in live computation and pre-computation scenarios. We avoid computationally intensive algorithms like LKH and CONCORDE due to the dynamic nature of mining environments, which require the flexibility to re-route the robot as needed in a limited amount of time which cannot be done with brute force algorithms.Our experiment does not intent to compare the various algorithms proposed in this study with the literature review as the dataset generated is very unique to the environment were the robot operates. 

    \item \textbf{First-of-its-Kind Review Research}: A significant contribution of our work is providing the first research review that compares the efficiency of various GOT optimization algorithms in solving real-world TSP problems. We critically evaluate these against basic ML optimization techniques, using real-world GPS data and benchmarking against famous pre-defined data like Concorde.
    
\end{enumerate}

This paper unfolds as follows: Section II reviews previous studies that have applied Google OR-Tools and reinforcement learning to the TSP. Section III provides a comprehensive overview of the various optimization techniques for TSP resolution. Section IV outlines our experimental design and methodology as well as the analyze of our experimental findings, alongside a depiction of the software interface where GPP is implemented. Lastly, Section V encapsulates the conclusions drawn from our research and hints at prospective directions for future studies.

\section{Related Works}

For TSPs, numerous methods have been employed to identify the most cost-efficient route. Comparative studies often benchmark their outcomes against exact algorithms, while computationally expensive to deliver optimal TSP tours. Prominently, the Concorde solver \cite{concordesolver} has been acclaimed as the most efficient, solving a TSP with 85,900 Euclidean cities \cite{concorde}. Given the substantial computational time exact algorithms require, alternative methods have been sought to approximate the exact solution with a tolerable deviation percentage. These algorithms, often classified as heuristics or metaheuristics, are favored for their computational efficiency and precision.\break

The Christofides Algorithm, developed in 1976 \cite{christo}, has proven to be a significant contribution to TSP solutions over the past decades. Empirical evaluations typically indicate that it achieves an accuracy of roughly 90\% when compared to an optimal TSP solution, thus providing a minimum performance guarantee of 50\%. An estimated improvement of \(1.5*10^{-36} \) was conducted by the following study \cite{improvedC}.\break

Various Heuristic methods have been compared over the years. A. Hanig Halim, et al., they compared six meta-heuristic algorithms (Nearest Neighbor, Genetic Algorithm, Simulated Annealing, Tabu Search, Ant Colony Optimization, Tree Physiology Optimization) on their ability to solve the TSP. Nearest Neighbor, Tree Physiology Optimization, and Genetic Algorithm were the fastest in computation, while Tabu Search and Ant Colony Optimization required more computation time. Tabu Search, Tree Physiology Optimization, and Genetic Algorithm achieved results nearest to the optimal solution, but the accuracy of Tabu Search declined with larger problems. The study highlights the need for more research in real-world contexts to better understand these algorithms' strengths and weaknesses.\break

A pertinent study aligned with ROMIE's application was conducted by Jie Chen, et al., where they addressed the TSP for unmanned aerial vehicles (UAVs) utilizing two parallel optimization algorithms. The study focuses on the application of two parallel optimization algorithms, Iterative Genetic Algorithm (IGA) and Particle Swarm Optimization - Ant Colony Optimization (PSO-ACO), for effective UAV path planning. It establishes a suitable TSP model for UAV path planning and demonstrates that these proposed algorithms provide more reasonable and effective solutions than comparative approaches, thereby optimizing UAV path planning \cite{UAV}.\break

Regrettably, as far as our research indicates, a comprehensive comparison among Google OR-Tools algorithms is absent from the literature. A single study \cite{comparativeGOOGLE} was found that made a comparison between heuristic methods from Google OR-Tools, ML, and Exact algorithms. Nevertheless, gaining insights into the behaviors of the aforementioned heuristics contributes to our understanding of the performance of Google OR-Tools, especially considering the similarity of some of their methods to the algorithms presented in the study by A. Hanig Halim, et al.\break

In the recent years, ML methods particularly Deep Neural Networks (DNNs) and RL, are revolutionizing the algorithmic landscape. Neural Networks (NN) have showcased potential in learning combinatorial algorithms, including both supervised and RL techniques.\break

This study \cite{metb} introduces 'Tspformer', a memory-efficient Transformer-based network model designed to address large-scale TSP issues. Tspformer incorporates a sampled scaled dot-product attention mechanism, significantly reducing time and space complexity. The model outperforms existing methods, handling up to 1000 city nodes, showing reduced memory usage and training time. Its effectiveness is validated against other models, demonstrating its potential as a new solution for TSP combinatorial optimization problems. \break

This paper \cite{ltsp} presents an end-to-end neural combinatorial optimization pipeline for the TSP. The study uncovers that learning scale-invariant TSP solvers requires a reassessment of neural combinatorial optimization, particularly regarding generalization. Key findings include the need for explicit redesigns to accommodate shifting graph distributions and the advantage of autoregressive decoding for generalization. The paper also highlights that models trained with expert supervision are more conducive to post-hoc search, whereas reinforcement learning scales better with increased computation. \break

The study \cite{htsp} introduces H-TSP, an end-to-end framework based on hierarchical reinforcement learning, tailored for large-scale TSP instances. H-TSP employs a two-tier policy structure, where the upper-level policy selects a subset of nodes, and the lower-level policy generates a tour connecting these nodes. The model demonstrates substantial efficiency, scaling to TSP instances of up to 10000 nodes, and offers comparable solution quality with significant time reduction compared to state-of-the-art (SOTA) search-based approaches.\break

DNNs can potentially enhance these heuristic methods. Training these networks might lead to more efficient algorithms for combinatorial optimization problems, posing the question if Deep Learning can supersede traditional heuristic algorithms \cite{DNNtsp}. \break

The study \cite{min2023unsupervised} proposes UTSP (Unsupervised Traveling Salesman Problem), a state-of-the-art unsupervised learning approach for solving the TSP using Graph Neural Networks (GNNs) with a surrogate loss function to overcome challenges faced by RL. The approach involves building a heatmap from GNN output and feeding it into a search algorithm and emphasizes the importance of expressive GNNs for efficiency. UTSP outperforms or competes with other learning-based TSP heuristics in terms of solution quality, running speed, and reduced training requirements.\break

Residual E-GAT \cite{Lei2022} showcases a DRL framework that employs an improved graph attention network (GAT) encoder paired with a Transformer decoder. Two deep reinforcement learning algorithms, Proximal Policy Optimization (PPO) and an enhanced REINFORCE algorithm, are utilized for model training, showing that PPO has superior sample utilization efficiency. Moreover, the framework exhibits linear time complexity during training and testing and generalizes well from training on random instances to testing on real-world problems, even with a smaller training dataset. \break

The MRAM method \cite{MRAM} introduces advanced node embeddings through batch normalization reordering and gate aggregation, and it produces dynamic-aware context embeddings using an attentive aggregation module on multiple relational structures. Experimentation across various VRP types, including the TSP and capacitated VRP (CVRP), shows the model's superiority over learning-based baselines and traditional methods, highlighting its speed and generalizability in larger-scale and varied distribution problems.\break

The G-DGANet method \cite{G-DGANet} is the latest DRL framework which presents a Gated Deep Graph Attention Network. The approach incorporates edge embedding and a gating mechanism within the framework, enhancing node learning and feature propagation. A graph pooling module aggregates node embeddings, leveraging a gate mechanism to capture the graph's global structure. The encoder-decoder architecture processes graph features and predicts node visitation sequences. The model is trained using proximal policy optimization, a reinforcement learning technique, to optimize the node selection policy for TSP solutions.\break

Numerous academic papers have scrutinized the continuously evolving state-of-the-art ML solutions for the TSP, including \cite{joshi2019efficient}  and \cite{kool2019attention}. These works provide extensive comparisons of ML models tested on varying numbers of cities (e.g., TSP20, TSP50, TSP100, TSP200, TSP500, and TSP1000). Notable algorithms like Concorde, Gurobi, and LKH3 generate optimal TSP solutions, with others applying ML to simple TSP heuristics or optimized problems. The analysis involves metrics, such as tour length, optimality gap, evaluation time, and the type of NN used by different deep learning approaches, including heuristic (H), supervised learning (SL), reinforcement learning (RL), sampling(s), greedy (G), beam search (BS), beam search and shortest tour heuristic (BS*), and 2OPT local search. \break


\begin{table}[h!]
    \caption{Comparison of SOTA ML models used to solve the TSP in contrast to well-known exact algorithms. The information for this table was compiled from various papers referenced earlier on the same Concorde dataset \cite{concorde}. TL represents Tour Length, G is the Gap in percentage to the Concorde solution, T is the computational Time.}
\centering
\small
\begin{tabular}{|l|l|c|c|c|c|c|c|c|c|c|}
\hline
\multicolumn{2}{|c|}{\textbf{Method}} & \multicolumn{3}{c|}{\textbf{TSP20}} & \multicolumn{3}{c|}{\textbf{TSP50}} & \multicolumn{3}{c|}{\textbf{TSP100}} \\ \hline
\textbf{} & \textbf{} & \textbf{TL} & \textbf{G} & \textbf{T} & \textbf{TL} & \textbf{G} & \textbf{T} & \textbf{TL} & \textbf{G} & \textbf{T} \\ \hline
\multirow{3}{*}{\begin{sideways}\textbf{Exact}\end{sideways}} & Concorde \cite{concordesolver} & 3.83 & 0.00 & 1m & 5.70 & 0.00 & 2m & 7.76 & 0.00 & 3m \\ \cline{2-11} 
 & LKH3 \cite{lkh3} & 3.83 & 0.00 & 18s & 5.70 & 0.00 & 5m & 7.76 & 0.00 & 21m \\ \cline{2-11} 
 & Gurobi \cite{gurobi} & 3.83 & 0.00 & 7s & 5.70 & 0.00 & 2m & 7.76 & 0.00 & 17m \\ \hline
\multirow{11}{*}{\begin{sideways}\textbf{Greedy}\end{sideways}} & Nearest Insertion & 4.33 & 12.91 & 1s & 6.78 & 19.03 & 2s & 9.46 & 21.82 & 6s \\ \cline{2-11} 
 & Random Insertion \cite{simple_heursitic} & 4.0 & 4.3 & 0s & 6.1 & 7.6 & 1s & 8.5 & 9.6 & 3s \\ \cline{2-11} 
 & Farthest Insertion \cite{simple_heursitic} & 3.9 & 2.3 & 1s & 6.0 & 5.5 & 2s & 8.3 & 7.5 & 7s \\ \cline{2-11} 
 & Nearest Neighbor \cite{simple_heursitic}& 4.5 & 17.2 & 0s & 7.0 & 22.9 & 0s & 9.6 & 24.7 & 0s \\ \cline{2-11} 
 & PtNet \cite{PtNet1} & 3.8 & 1.1 & NA & 7.6 & 34.4 & NA & NA & NA & NA \\ \cline{2-11} 
 & PtNet \cite{PtNet2} & 3.8 & 1.4 & NA & 5.9 & 4.4 & NA & 8.3 & 6.9 & NA \\ \cline{2-11} 
 & S2V \cite{S2V} & 3.8 & 1.4 & NA & 5.9 & 5.1 & NA & 8.3 & 7.0 & NA \\ \cline{2-11} 
 & GAT \cite{GAT1} & 3.8 & 0.6 & 2m & 5.9 & 3.9 & 5m & 8.4 & 8.4 & 8m \\ \cline{2-11} 
 & GAT \cite{GAT1} & 3.8 & 0.4 & 4m & 5.8 & 2.7 & 26m & 8.1 & 5.2 & 3h \\ \cline{2-11} 
 & GAT \cite{GAT2} & 3.8 & 0.3 & 0s & 5.8 & 1.7 & 2s & 8.1 & 4.5 & 6s \\ \cline{2-11} 
 & GCN \cite{GCN} & 3.8 & 0.6 & 6s & 5.8 & 3.1 & 55s & 8.4 & 8.3 & 6m \\ \hline
\multirow{19}{*}{\begin{sideways}\textbf{Sampling/Heuristic Search}\end{sideways}} & OR Tools \cite{ortools} & 3.8 & 0.3 & NA & 5.8 & 1.8 & NA & 7.9 & 2.9 & NA \\ \cline{2-11} 
 & Chr.f + 2OPT \cite{NNnaive} & 3.8 & 0.3 & NA & 5.7 & 1.6 & NA & NA & NA & NA \\ \cline{2-11} 
 & Genetic Algorithms \cite{GAalgo} & NA & NA & NA & NA & NA & NA & NA & NA & NA \\ \cline{2-11} 
 & GNN \cite{GNN} & 3.9 & 2.4 & NA & NA & NA & NA & NA & NA & NA \\ \cline{2-11} 
 & PtNet \cite{PtNet2} & NA & NA & NA & 5.7 & 0.9 & NA & 8.0 & 3.0 & NA \\ \cline{2-11} 
 & GAT \cite{GAT1} & 3.8 & 0.1 & 5m & 5.7 & 1.2 & 17m & 8.7 & 12.7 & 56m \\ \cline{2-11} 
 & GAT \cite{GAT1} & 3.8 & 0.2 & 2s & 5.7 & 1.2 & 1h & 8.1 & 3.7 & 4.5m \\ \cline{2-11} 
 & GAT \cite{GAT2}  & 3.8 & 0.1 & 5m & 5.7 & 0.5 & 24m & 7.9 & 2.3 & 1h \\ \cline{2-11} 
 & GAT \cite{GAT2}& 3.8 & 0.2 & 6s & 5.7 & 1.6 & 35s & 8.1 & 4.3 & 1.8m \\ \cline{2-11} 
& GAT \cite{GAT2}& 3.8 & 0.2 & 6s & 5.7 & 1.6 & 35s & 8.1 & 4.3 & 1.8m \\ \cline{2-11} 
 & GCN \cite{GCN} & 3.8 & 0.01 & 15m & 5.7 & 0.2 & 26m & 7.9 & 2.4 & 1.7h \\ \cline{2-11} 
 & GCN \cite{GCN}& 3.8 & 0.6 & 20s & 5.8 & 3.5 & 2m & 8.4 & 8.3 & 11m \\ \cline{2-11} 
 & GCN \cite{GCN}& 3.8 & 0.0 & 12m & 5.7 & 0.1 & 18m & 7.8 & 1.3 & 40m \\ \cline{2-11} 
& Att-GCRN \cite{GCN} & 3.8 & -0.01 & 1m & 5.6 & 0.0 & 6 & 7.7 & 0.0 & 9m \\ \cline{2-11} 
 & MvRAM \cite{MRAM} & 3.8 & 0.2 & 0.3a & 5.7 & 1.2 & 1s & 8.1 & 3.7 & 2s \\ \cline{2-11} 
 & Residual RL+GAT \cite{Lei2022} & 3.8 & 0.1 & 10m & 5.7 & 0.7 & 55m & 7.8 & 1.7 & 2h \\ \cline{2-11} 
 & UTSP \cite{min2023unsupervised} & 3.8 & -0.0 & 1m & 5.6 & -0.01 & 2.8m & 7.7 & -0.01 & 11m \\ \cline{2-11} 
 & G-DGANet \cite{G-DGANet} & 3.8 & 0.1 & 12m & 5.7 & 0.4 & 1h & 7.7 & 0.4 & 2.1h \\ \hline
\end{tabular}
\end{table}

\begin{table}[h!]
    \caption{Comparison of SOTA ML models used to solve the TSP in contrast to well-known exact algorithms. The information for this table was compiled from various papers referenced earlier on the same Concorde dataset \cite{concorde}. TL represents Tour Length, G is the Gap in percentage to the Concorde solution, T is the computational Time.}
\centering
\small
\begin{tabular}{|l|l|c|c|c|c|c|c|c|c|c|}
\hline
\multicolumn{2}{|c|}{\textbf{Method}} & \multicolumn{3}{c|}{\textbf{TSP200}} & \multicolumn{3}{c|}{\textbf{TSP500}} & \multicolumn{3}{c|}{\textbf{TSP1000}} \\ \hline
\textbf{} & \textbf{} & \textbf{TL} & \textbf{G} & \textbf{T} & \textbf{TL} & \textbf{G} & \textbf{T} & \textbf{TL} & \textbf{G} & \textbf{T} \\ \hline
\multirow{3}{*}{\begin{sideways}\textbf{Exact}\end{sideways}} & Concorde & 10.72 & 0.00 & 3.4m & 16.55 & 0.00 & 38m & 23.12 & 0.00 & 6.7h \\ \cline{2-11} 
 & LKH3 \cite{lkh3} & 10.70 & -0.14 & 41m & 16.52 & -0.17 & 47m & NA & NA & NA \\ \cline{2-11} 
 & Gurobi \cite{gurobi} & 10.72 & 0.00 & 2m & 16.55 & 0.00 & 11.5m & 23.12 & 0.00 & 38m \\ \hline
\multirow{11}{*}{\begin{sideways}\textbf{Greedy}\end{sideways}} && NA & NA & NA & NA & NA & NA & NA & NA & NA \\ \cline{2-11} 
 & Random Insertion \cite{simple_heursitic} & NA & NA & NA & NA & NA & NA & NA & NA & NA \\ \cline{2-11} 
 & Farthest Insertion \cite{simple_heursitic} & NA & NA & NA & NA & NA & NA & NA & NA & NA \\ \cline{2-11} 
 & Nearest Neighbor \cite{simple_heursitic} & NA & NA & NA & NA & NA & NA & NA & NA & NA \\ \cline{2-11} 
 & PtNet \cite{PtNet1} & NA & NA & NA & NA & NA & NA & NA & NA & NA \\ \cline{2-11} 
 & PtNet \cite{PtNet2} & NA & NA & NA & NA & NA & NA & NA & NA & NA \\ \cline{2-11} 
 & S2V \cite{S2V} & NA & NA & NA & NA & NA & NA & NA & NA & NA\\ \cline{2-11} 
 & GAT \cite{GAT1} & NA & NA & NA & NA & NA & NA & NA & NA & NA \\ \cline{2-11} 
 & GAT \cite{GAT1} & NA & NA & NA & NA & NA & NA & NA & NA & NA \\ \cline{2-11} 
 & GAT \cite{GAT2} & NA & NA & NA & NA & NA & NA & NA & NA & NA \\ \cline{2-11} 
 & GCN \cite{GCN} & NA & NA & NA & NA & NA & NA & NA & NA & NA \\ \hline
\multirow{19}{*}{\begin{sideways}\textbf{Sampling/Heuristic Search}\end{sideways}} & OR Tools \cite{ortools} & NA & NA & NA & NA & NA & NA & NA & NA & NA \\ \cline{2-11} 
 & Chr.f + 2OPT \cite{NNnaive} & NA & NA & NA & NA & NA & NA & NA & NA & NA \\ \cline{2-11} 
 & Genetic Algorithms \cite{GAalgo} & NA & NA & NA & NA & NA & NA & NA & NA & NA \\ \cline{2-11} 
 & GNN \cite{GNN} & NA & NA & NA & NA & NA & NA & NA & NA & NA \\ \cline{2-11} 
 & PtNet \cite{PtNet2} & NA & NA & NA & NA & NA & NA & NA & NA & NA \\ \cline{2-11} 
 & GAT \cite{GAT1} & 13.17	& 22.91	& 5m	& 28.63 &	73.03	& 21m &	50.30	& 117.59	&37m \\ \cline{2-11} 
 & GAT \cite{GAT1} & 11.61&	8.32&	10m&	23.75&	43.57&	1h&	47.73&	106.46&	5.4h \\ \cline{2-11} 
 & GAT \cite{GAT2} & 11.45&	6.82	&4.5m	&22.64	&36.84	&16m&	42.80&	85.15&	64m\\ \cline{2-11} 
 & GAT \cite{GAT2} & 11.61	&8.31&	5m	&20.02	&20.99&	2m	&31.15	&34.75	&3m \\ \cline{2-11} 
 & GAT \cite{GAT2} & 11.38&	6.14&	6m	&19.53	&18.03	&22m&	29.90&	29.24&	1.6h \\ \cline{2-11} 
 & GCN \cite{GCN} & 17.01&	58.73&	1m	&29.72	&79.61&	7m	&48.62	&110.29	&29m \\ \cline{2-11} 
 & GCN \cite{GCN} & 16.19&	51.02&	4.6m	&30.37&	83.55&	38m	&51.26	&121.73&	52m \\ \cline{2-11} 
 & GCN \cite{GCN} & 16.21&	51.21&	4m	&30.43	&83.89	&31m	&51.10&	121.04&	3.2h \\ \cline{2-11} 
& Att-GCRN \cite{GCN} & 10.74 & 0.16 & 1.5m & 16.75 & 1.22 & 4m & 23.52 & 1.72 & 8m \\ \cline{2-11} 
 & MvRAM \cite{MRAM} & NA & NA & NA & NA & NA & NA & NA & NA & NA \\ \cline{2-11} 
 & Residual RL+GAT \cite{Lei2022} & NA & NA & NA & NA & NA & NA & NA & NA & NA \\ \cline{2-11} 
 & UTSP \cite{min2023unsupervised} & 10.73 & 0.09 & 2m & 16.68 & 0.84 & 3m & 23.39 & 1.18 & 5m \\ \cline{2-11} 
 & G-DGANet \cite{G-DGANet} & NA & NA & NA & NA & NA & NA & NA & NA & NA \\ \hline
\end{tabular}
\end{table}


A RL-driven Iterated Greedy Algorithm (RLIGA) for the TSP is proposed in \cite{RLIGA}. A key component is the incorporation of a Q-Learning which guides the algorithm to select the optimal parameters for iterative searches based on historical experience with a unique aspect of this approach being the introduction of a 'damage size $d$' selection mechanism informed by Q-learning. This mechanism is designed to prevent the algorithm from engaging in blind searches, increasing its ability to find optimal solutions. \break

Studies from Jiazhao Liang \cite{TSPQL} suggest that the potential use of the Q-learning and RL method to solve the TSP. The approach, while currently based on a simple model focused on distance, demonstrates successful convergence. The study indicates the need for future work to incorporate more complex factors for a better reality model. The authors demonstrate the proposed Q-learning method in the case of optimal path finding among warehouses, underscoring the relevance of RL in addressing combinatorial optimization problems. \break

Wang, et al. \cite{comparisonML} applied three RL algorithms, such as Q-Learning, Sarsa, and Double Q-Learning, to solve the TSP. They found the Double Q-Learning algorithm to be the most effective, when using a reward function of $R1$ = 1/$dij$. \break

Despite the growing utilization of optimization algorithms like Google OR-Tools and Q-Learning methods in tackling complex challenges such as the TSP, there remains a notable void in the literature: a thorough performance analysis and comparison of these techniques, particularly in the context of UGV applications. Existing studies have not fully explored the application of these optimization tools in UGV scenarios, nor have they conducted a comprehensive comparison of all available optimization algorithms in relation to the TSP. This gap in the research limits our understanding of the relative strengths and weaknesses of these diverse approaches when they are applied to practical, real-world challenges. Our study addresses this gap by providing an unprecedented comparative analysis between Google OR-Tools optimization algorithms and Q-Learning methods. We aim to evaluate their effectiveness specifically in optimizing TSP solutions, with a unique focus on their application in GPP and an exhaustive comparison across the spectrum of available TSP optimization algorithms. This approach sets our research apart and contributes significantly to the field by illuminating the potential of these methods in both theoretical and practical UGV applications.

\section{Optimization Methods for solving the Traveling Salesman Problem}
\label{Optimization Methods for solving the Traveling Salesman Problem}

    \subsection{Traveling Salesman Problem}

The TSP is a renowned NP-hard problem \cite{attention} in the field of combinatorial optimization, which poses a significant query: ”Given n cities and the distances between each pair, what is the most efficient route that visits each city once and returns
to the starting city?” \cite{efficient}. This problem, pivotal in industrial sectors, such as transportation becomes exponentially complex when the number of nodes or cities increases, thereby underscoring the necessity for effective resolutions. \break

A mathematical interpretation of the problem would involve grasping the vast number of potential solutions that route optimization can generate. There are $(n-1)!/2$ possible combinations. \cite{lake} Additionally, the researchers can depict the problem through different methods to present the optimal path. Assuming $P_n$ as the total permutations of the set {1,2,\ldots,$n$} and $n$ equals $x$ points. The TSP involves the quest for $\pi$ = $(\pi(1), \pi(2), \ldots, \pi(n))$ in $P_n$ so that the function in $Eq.1$  is minimized. Here, $C$ signifies the cost or distance between two cities and is known as the tour length:

            \begin{equation}
                C_{\pi(n)}\pi(1) + \sum_{i=1}^{n-1} C_{\pi(i)\pi(i+1)}
            \end{equation}

Because the exact solution is expensive to find, a plethora of meta-heuristic algorithms, each embodying unique strategies of intensification and diversification, have been put forward in scholarly literature to seek gratifying solutions for the TSP \cite{lake}. ML [4] offers an alternative route to crafting solutions manually, which could be costly
or demand a high level of specialized knowledge. Recent strides in graph neural network
techniques stand out as they inherently function on the graph structure of such problems.
Thus, the creation and evaluation of algorithms for resolving the TSP remains a vibrant area of investigation in the research community. 

    \subsection{Google OR-Tools}

Google’s OR-Tools are an open-source suite designed for optimization, providing a plethora of algorithms for routing, flow, graph, linear programming, and constraint problems. In the context of the TSP, OR-Tools offer a robust and adaptable platform for constructing and solving TSP instances \cite{ortools}. The library includes a variety of first solution and local search algorithms, forming a comprehensive online toolset for addressing the TSP. Table \ref{table:1} below highlights these useful algorithms provided by the OR-Tools API.

\begin{table}[ht]
\caption{Various local search and meta-heuristic methods used in OR-Tools.}
\label{table:1}
\centering
\begin{tabular}{|c|c|c|}
\hline
\textbf{Category} & \textbf{Method} & \textbf{Algorithm}\\
\hline
\multirow{4}{*}{First Local Search} & PATH\_CHEAPEST\_ARC (PCA) & Greedy \\
\cline{2-3}
 & PATH\_MOST\_CONSTRAINED\_ARC (PMCA) & Greedy Constrained-based \\
 & LOCAL\_CHEAPEST\_INSERTION (LCI) & Local insertion heuristic \\
\cline{2-3}
 & GLOBAL\_CHEAPEST\_ARC (GCA) & Greedy \\
 & LOCAL\_CHEAPEST\_ARC (LCA) & Localised Greedy \\
\cline{2-3}
 & FIRST\_UNBOUND\_MIN\_VALUE (FUMV) & Value assignment \\
 & SAVINGS(s) & Savings algorithm (Clarke \& Wright) \cite{saving} \\
 & CHRISTOFIDES (C) & Graph Theory heuristic \\
\cline{2-3}
 & PARALLEL\_CHEAPEST\_INSERTION (PCI) & Heuristic \\
\hline
\multirow{4}{*}{Meta-heuristic} & GREEDY\_DESCENT (GD) (Hill Climbing) & Greedy heuristic \\
\cline{2-3}
 & GUIDED\_LOCAL\_SEARCH (GLS) & Penalty-based \\
\cline{2-3}
 & SIMULATED\_ANNEALING (SA) & Probabilistic \\
\cline{2-3}
 & TABU\_SEARCH (TS) & Memory-based \\
 & GENERIC\_TABU\_SEARCH (GTS) & Memory-based \\
\hline
\end{tabular}
\end{table}

A myriad of renowned algorithms for resolving the TSP, as shown in Table \ref{table:1}, have been explored. Google OR-Tools provides a swift avenue to experiment with TSP solutions, offering pre-implemented optimization algorithms through its API. However, it should be noted that this approach restricts full control over their implementation details.

There are some high level explanations of the theory behind some of the deterministic and stochastic algorithms \cite{deterministic}.

\begin{itemize}
    \item \textbf{Nearest Neighbour (Path\_Cheapest\_Arc)}

    The Nearest Neighbor (NN) algorithm is a commonly used heuristic for the TSP. It starts from a random city and sequentially visits the nearest unvisited city until all cities are visited. Despite its non-optimal nature \cite{wikiNN}, the NN algorithm is efficient and practical with a runtime complexity of O(n²), where n is the number of cities. It offers rapid solutions for small to moderately large problem instances \cite{NNnaive}.

    \item \textbf{Minimum Spanning Tree (Similar to Local\_Cheapest\_Insertion)}

The Minimum Spanning Tree (MST) Heuristic, used in solving the TSP, finds the minimum total edge cost that connects all nodes in a graph. The TSP constraints are met by adding more edges to the MST (Fig.~\ref{fig:mstcost1}), ensuring each node is visited once without backtracking \cite{mstwiki}.

\begin{figure}[H]
\centering
\subfloat[MST tour]{
  \includegraphics[width=0.4\textwidth]{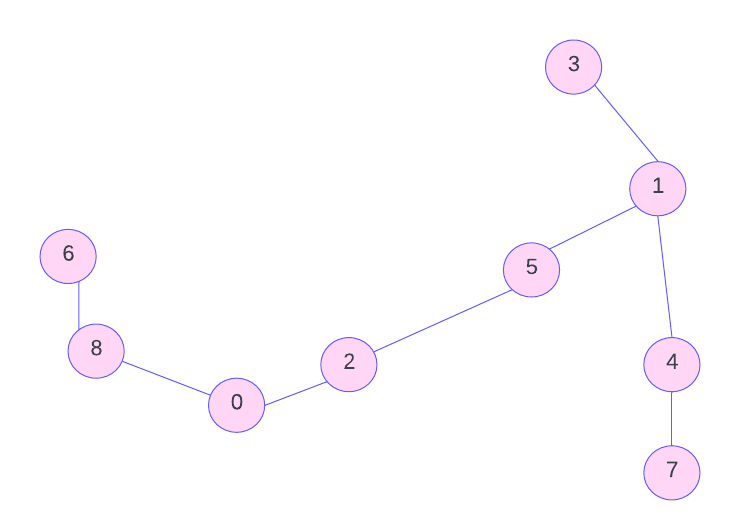}
  \label{fig:MST tour}
}
\qquad
\subfloat[optimal TSP tour]{
  \includegraphics[width=0.4\textwidth]{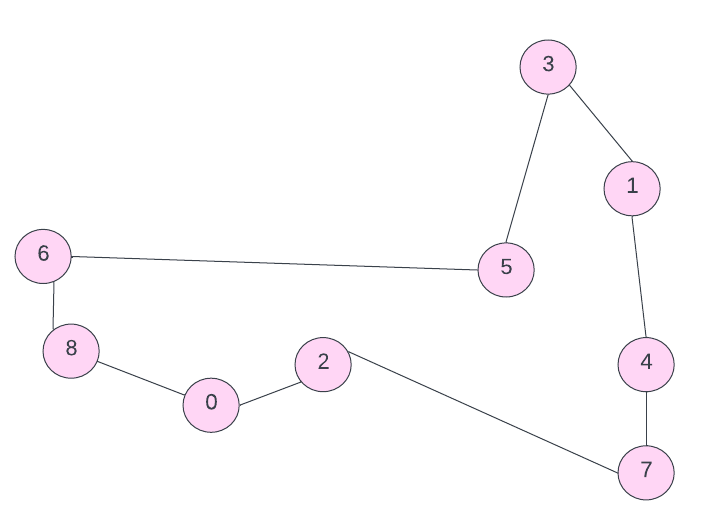}
  \label{fig:Optimal TSP tour}
}
\caption{MST cost compared to optimal TSP with MST cost inferior to the Optimal TSP tour}
\label{fig:mstcost1}
\end{figure}


The proposed MST implementation employs Prim’s Algorithm, which has a time complexity of $O(|E| + |n|log(|n|))$, where $|E|$ represents the edge number and $|n|$ the vertex number. In TSP’s complete graph context, $|E|$ simplifies to $|n|(|n| - 1)$ \cite{NNnaive}. The MST heuristic builds a foundation for TSP heuristic algorithms by creating an MST, doubling each edge, finding an Eulerian circuit in the doubled tree, and finally shortcutting the circuit into a Hamiltonian cycle to form a TSP tour. While not always yielding the optimal TSP solution, the MST heuristic offers a valuable starting point or approximation method, especially when combined with Prim’s or Kruskal’s Algorithm. The MST's cost will always be less than the TSP's cost, offering a lower bound for the TSP and a means to conceptualize the optimal TSP tour by creating a "\textbf{1-tree}" through edge removal (Fig.~\ref{fig:mst1tree2}). To find the 1-tree, we first remove any vertex v and find the MST has described above. Secondly, we connect two shortest edges to the vertex. At the end 1-tree cost is inferior to TSP cost.

\begin{figure}[H]
\centering
\subfloat[1-tree tour]{
  \includegraphics[width=0.4\textwidth]{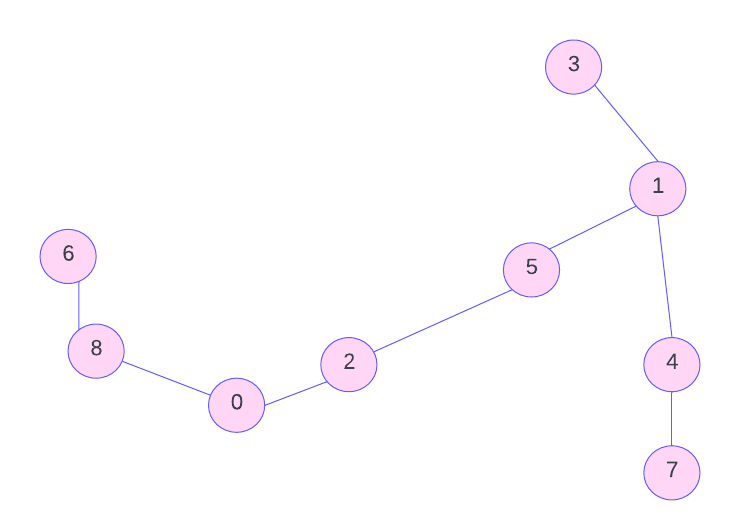}
  \label{fig:1-tree tour}
}
\qquad
\subfloat[optimal TSP tour]{
  \includegraphics[width=0.4\textwidth]{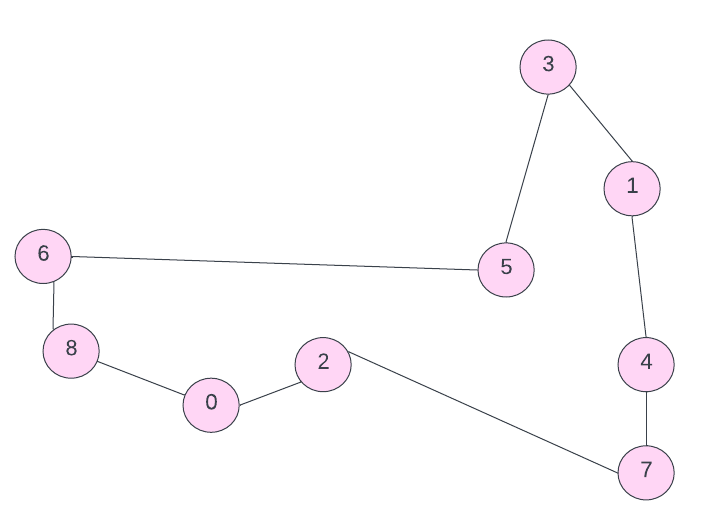}
  \label{fig:Optimal TSP tour_1tree}
}
\caption{MST 1-tree with improved lower bound.}
\label{fig:mst1tree2}
\end{figure}


    \item \textbf{Christofides}: 
    
The Christofides' Algorithm, a renowned method in theoretical Computer Science, for TSP. This multistep algorithm guarantees a solution within 1.5 times the optimal. This algorithm encompasses several steps:

\begin{enumerate}
    \item \textbf{MST}: Develop a Minimum Spanning Tree.
    \item \textbf{Odd degrees}: Identify vertices with odd degrees in MST, ensuring no node repetition. This step, with a time complexity of $O(|n|)$, relies on available degree count for each vertex.
    \item \textbf{Matching}: Find smallest distance matchings, potentially using algorithms like Blossom V. This step, the most computationally expensive, has a runtime complexity of at least $O(|n|^4)$ \cite{blossom}.
    \item \textbf{Euler}: Combine matching and MST to form an Eulerian multigraph \cite{euler}, ensuring even node degrees. This multigraph underlies the Eulerian tour, computed using Fleury’s algorithm \cite{fleury}.
    \item \textbf{Hamiltonian}: Verify that each city has been visited; if yes, proceed to the next city (Fig.~\ref{fig:christo}).
\end{enumerate}

\begin{figure}[H]
\centering
\includegraphics[width=1\columnwidth, height=0.85\columnwidth, keepaspectratio]{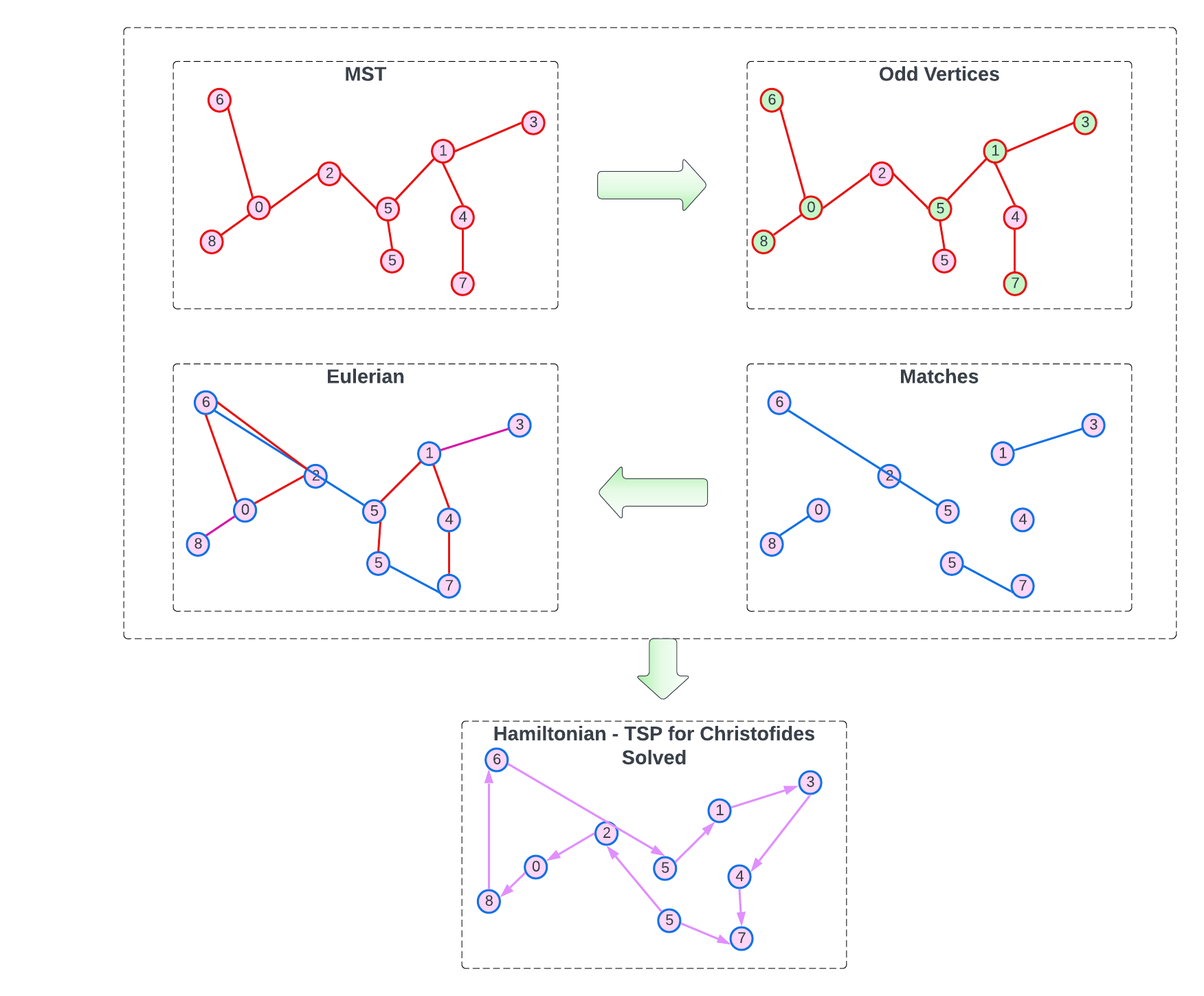}
\caption{Christofides step by step graphical representation.}
\label{fig:christo}
\end{figure}

    \item \textbf{Hill Climbing}

Hill Climbing begins with a random sequence of cities and iteratively improves it. The algorithm considers all successor states, each obtained by swapping two adjacent cities, and chooses the best one as the new state. This process continues until an adequate solution is found. Although Hill Climbing is simple and efficient, it's limited by its short-sighted approach, often stopping at local optima and potentially missing the global optimum \cite{HC}.
    
    \item \textbf{Simulated Annealing}

The Simulated Annealing algorithm is a probabilistic search technique that aims to find an optimal solution by exploring the solution space, allowing for the acceptance of worse solutions to escape local optima. It uses a temperature ($T$) parameter that gradually decreases over time, controlling the probability of accepting worse solutions. The cooling schedule, represented by the parameter $\alpha$, determines how the temperature decreases.

A mathematical representation of the probability of choosing changes in a solution (move) is calculated as:

                    \begin{equation}
                        P(\text{move A → B}) = e^{-\Delta/T}
                    \end{equation}

                where $\delta = f(A) - f(B)$ and $T$ is the current temperature. The temperature is set to decrease by \cite{halim}:

\begin{equation}
    T = \alpha T_0
\end{equation}

                where $\alpha$ is a number in (0,1) acting as a decreasing factor.

                The following figure describes the probabilistic acceptance that the SA method follows for exploring the solution space and helping in avoiding getting trapped in local optima \cite{ieeeSA}.
Decreasing the temperature helps with this transition. In the final stages of the search, the minimum number of bad solutions is to be chosen to improve the accuracy of the solution as much as possible. This is called exploitation. The only mechanism to
balance exploitation and exploration in the Simulated Annealing algorithm is the temperature.

    \item \textbf{Tabu Search}
    
Tabu Search, is a global optimization algorithm (see flowchart in Fig.~\ref{fig:tabufc}). Starting from an initial feasible solution, it explores the solution's neighborhood to identify candidate solutions. If a candidate fulfills predetermined rules, its tabu status is bypassed, it is deemed as the current solution and potentially the global optimum, and added to the tabu list. If not, the optimal solution among the non-taboo ones is selected as the current optimal solution, also added to the tabu list. This process continues until a termination condition is met \cite{tabuieee} \cite{tabuCS}.

\begin{figure}[H]
\centering
\includegraphics[width=0.7\columnwidth, height=0.85\columnwidth, keepaspectratio]{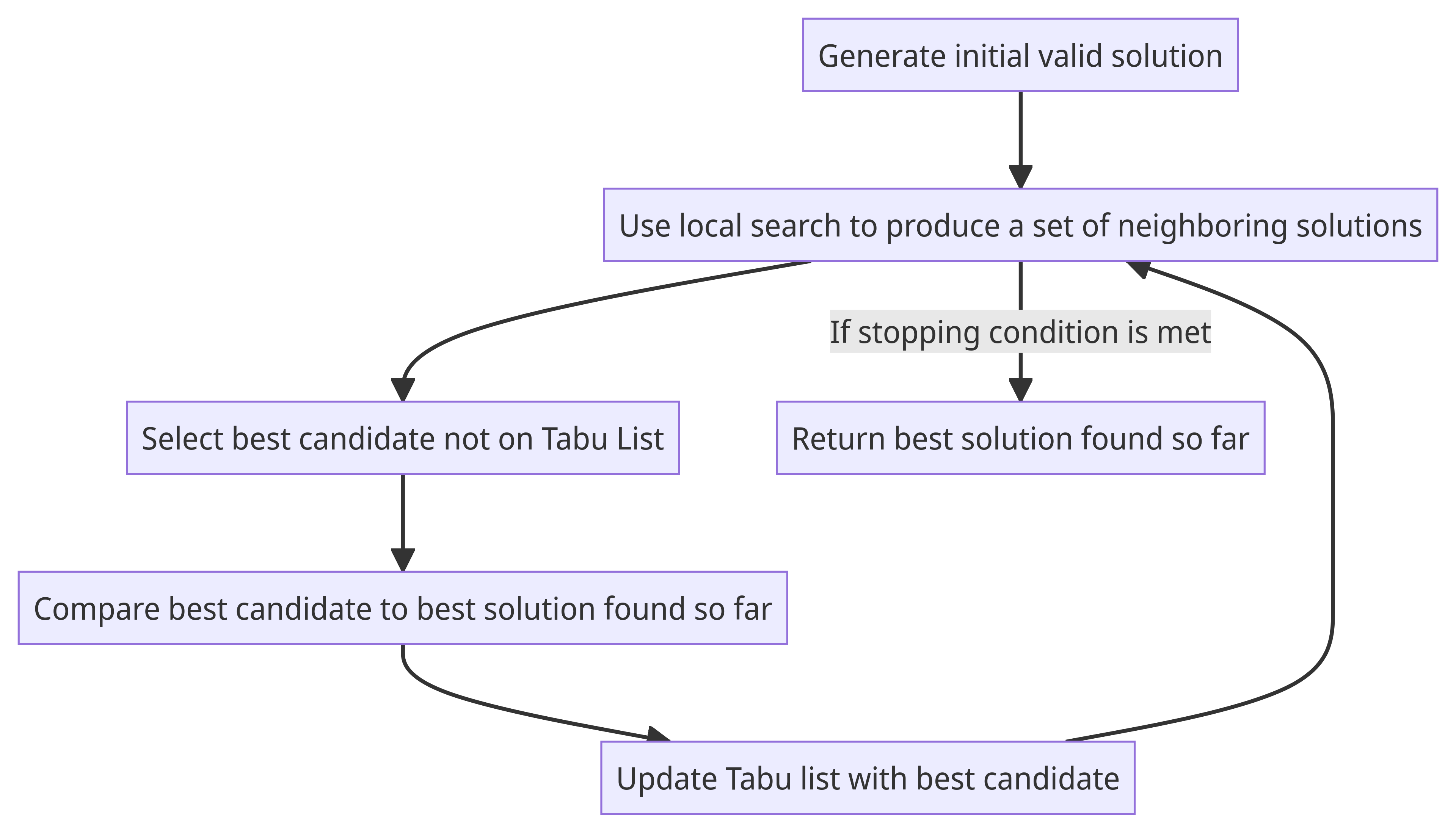}
\caption{Flowchart of Tabu search process.}
\label{fig:tabufc}
\end{figure}

\end{itemize}

    \subsection{RL: Reinforcement Learning}

Geochemical sampling processes may require more detailed prospection over time, necessitating scalability and responsiveness to customer needs in the field. In the event that the customer
wishes to sample a specific area following promising results on the sampling heat maps, the use of
ML algorithms, such as UTSP may provide a viable solution. However, the application of such algorithms can be complex and present certain limitations, including an internal lack of current knowledge of Graph Neural Networks (GNN) for optimal implementation. Nevertheless, some state-of-the-art solutions offer a reasonable degree of optimality and free resources are available, although scalability for high point counts remains an issue. Q-learning, a type of RL, is selected for its straightforward implementation and comprehensibility. It offers a practical and intuitive approach to solving problems, making it accessible and easy to understand. The alternative description of this method was not provided in the preceding section. \break

\par \textbf{\textit{1) Q-Learning}}: The Q-Learning is a model-free RL algorithm (Algorithm 1) \cite{article}. In this work, the Q-learning is applied to a 2D map. The Q-learning algorithm learns a policy to determine the optimal actions, i.e., moving from one point to another. The objective is to maximize the cumulative reward, i.e., minimizing the overall distances of the tour. The most important mathematical feature is the Q-Table which maps states and actions to Q-values. It estimates the quality of taking certain action in a given state. The Q-Values are represented as a state-action function ($s$,$a$). This pair is a cumulative reward starting in state ’$s$’ (sampling points) taking an action ’$a$’ (choice
of the next sampling points). It uses the Bellman equation as a simple value iteration update, using the weighted average of the current value and the new information \cite{qlearning}.

            \begin{equation}
                Q^{n}(s_{t}, a_{t}) = (1-\alpha)Q(s_{t}, a_{t}) + \alpha \left( r_{t} + \gamma \max_{a}Q(s_{t+1}, a) \right)
                \label{equat}
            \end{equation}

 \noindent where $r_{t}$ is the reward received when moving from state $s_{t}$ to state $s_{t+1}$, and $\alpha$ is the learning rate ($0 < \alpha \leq 1$). $Q^{new}(s_{t}, a_{t})$ is the updated Q value and it is composed of three factors:
            
            \begin{itemize}
                \item $(1-\alpha)Q(s_{t}, a_{t})$: the current value (weighted by one minus the learning rate)
                \item $\alpha r_{t}$: the reward $r_{t}$ to obtain if action $a_{t}$ is taken when in state $s_{t}$ (weighted by the learning rate)
                \item $\alpha \gamma \max_{a}Q(s_{t+1}, a)$: the maximum reward that can be obtained from state $s_{t+1}$ (weighted by the learning rate and discount factor $\gamma$)
            \end{itemize}

The learning process in Q-learning (Eq. 1) involves iteratively solving the TSP on the same set of points. After each step, the agent updates the Q-table by incorporating the received reward and estimating the Q-values of the next state. 

\begin{algorithm}
\caption{Q-Learning}
\begin{algorithmic}
\STATE Initialize Q-Table with zeros
\STATE Initialize $\epsilon$, $\epsilon_{\text{min}}$, $\epsilon_{\text{decay}}$, $\gamma$, $\alpha$
\FOR {each episode}
    \STATE Initialize state $s$
    \WHILE {episode is not done}
        \STATE Choose action $a$ from state $s$ using policy derived from Q-Table (e.g., $\epsilon$-greedy)
        \STATE Take action $a$, observe reward $r$, next state $s'$
        \STATE $Q[s,a] \leftarrow Q[s,a] + \alpha \cdot (r + \gamma \cdot \max_a Q[s',a] - Q[s,a])$
        \IF {$\epsilon > \epsilon_{\text{min}}$}
            \STATE $\epsilon \leftarrow \epsilon \cdot \epsilon_{\text{decay}}$
        \ENDIF
        \STATE $s \leftarrow s'$
    \ENDWHILE
\ENDFOR
\end{algorithmic}
\end{algorithm}

\break

\par \textbf{\textit{2) Double Q-Learning}}: The double Q-Learning (Algorithm 2) is an enhancement over Q-learning, designed to mitigate the overestimation issues in action value approximations inherent in Q-learning. It does this by using two separate Q-tables instead of one, each updated using information from the other, providing a more robust and unbiased estimate of action values which is different from the vanilla Q-Learning only taking the maximum Q-value for the updates. However, it can occasionally lead to underestimations and requires more computational resources due to the maintenance of two Q-tables. It is particularly beneficial in environments with high variance in the reward structure \cite{doubleql}.

    \begin{equation}
        Q_{t+1}^{A}(s, a) \&= Q_{t}^{A}(s, a) + \alpha \biggl[ r(s, a) + \gamma \max_{a'} Q_{}^{B}(s', a') - Q_{t}^{A}(s, a) \biggr]
        \label{eq:12ql}
    \end{equation}
    
    \begin{equation}
        Q_{t+1}^{B}(s, a) \&= Q_{t}^{B}(s, a) + \alpha \biggl[ r(s, a) + \gamma \max_{a'} Q_{}^{A}(s', a') - Q_{t}^{B}(s, a) \biggr]
        \label{eq:22ql}
    \end{equation}

The equation above is similar to the vanilla Q-Learning but the difference lies in how these tables are updated in this study. When the algorithm is about to update Q-TableA (QA), the best action (the action with the highest Q-value) is selected based on QA, but it uses the Q-value for that action from Q-TableA (QB) to perform the update. Similarly, when updating QB, the proposed algorithm selects the best action based on QB, but uses the Q-value from QA to perform the update.

\begin{algorithm}
\caption{Double Q-Learning}
\begin{algorithmic}
\STATE Initialize Q-TableA and Q-TableB with zeros
\STATE Initialize $\epsilon$, $\epsilon_{\text{min}}$, $\epsilon_{\text{decay}}$, $\gamma$, $\alpha$
\FOR {each episode}
    \STATE Initialize state $s$
    \WHILE {episode is not done}
        \STATE Choose action $a$ from state $s$ using policy derived from Q-Table (e.g., $\epsilon$-greedy)
        \STATE Take action $a$, observe reward $r$, next state $s'$
        \IF {$\text{rand()} < 0.5$}
            \STATE $a' \leftarrow \arg\max_a QA[s',a]$
            \STATE Update QA with Eq. 2
        \ELSE
            \STATE $a' \leftarrow \arg\max_a QB[s',a]$
            \STATE Update QB with Eq.3
        \ENDIF
        \IF {$\epsilon > \epsilon_{\text{min}}$}
            \STATE $\epsilon \leftarrow \epsilon \cdot \epsilon_{\text{decay}}$
        \ENDIF
        \STATE $s \leftarrow s'$
    \ENDWHILE
\ENDFOR
\end{algorithmic}
\end{algorithm}

\par \textbf{\textit{3) Epsilon greedy}}:

The epsilon-greedy strategy is a popular method for handling the exploration-exploitation dilemma in RL. The dilemma arises because, to find the optimal policy, an agent needs to explore its environment to learn the reward associated with various actions in different states. But, the agent also needs to exploit its current knowledge to maximize the rewards it receives. If it only exploits its current knowledge, it may never discover better policies. If it only explores, it will fail to maximize the rewards that it could have received based on its current knowledge.

\par The epsilon-greedy strategy is a simple method for balancing exploration and exploitation. It works as follows:

\begin{enumerate}
    \item With probability 1 - epsilon, the agent selects the action that it believes has the maximum expected reward. This is the exploitation part.
    \item With probability epsilon, the agent selects an action randomly. This is the exploration part.
\end{enumerate}

\par The parameter epsilon can be any value between 0 and 1, and it controls the balance between exploration and exploitation. A higher value means more exploration and less exploitation, while a lower value means more exploitation and less exploration \cite{egreedy}
\cite{egreedy2}.

\par The epsilon-greedy strategy can be formalized as:
\begin{equation*}
\begin{cases}
max_a Q_t(s,a), &{\text{with probability}}\ 1 - \epsilon,\\
{random,} &{\text{from all actions with probability }} \epsilon 
\end{cases}
\end{equation*}

\par \textbf{\textit{4) Model Definition}}

\begin{enumerate}
    \item \textbf{State}: In our context of TSP, a state is represented by the current position of the agent and the set of unvisited sampling points.
    \item \textbf{Action}: An action corresponds to moving from the current sampling point position to an unvisited city.
    \item \textbf{Reward}: The reward function in TSP could be designed to encourage shorter tours. A common approach is to use the negative of the travel cost as the reward. For example, if the agent moves from sampling point 1 to sampling 2 and the cost of travel is 10 units, then the reward would be -10. The goal of the agent (Algorithm 1) would be to maximize the total reward, which would equate to finding the shortest possible tour.
\end{enumerate}

\section{Results Discussion: The Difference Between Local and Global Optimum}

As elucidated in Part~\ref{Optimization Methods for solving the Traveling Salesman Problem}, certain metaheuristics have a propensity to get ensnared in local optima, engendering fluctuations in the cost function's behavior. A useful approach to comprehending the results of various Google OR-Tools algorithms is to anticipate the performance each algorithm might deliver. Commonly utilized algorithms like Hill Climbing (HC) and Simulated Annealing (SA) elucidate the issue of an agent becoming stuck or attempting to evade local optima. We have simulated a solitary peak landscape (Eq. 7) and a multiple peak landscape (Eq. 8), each represented as a 3D surface. The intent is to comprehend the agent's behavior in each scenario by assigning identical initial positions in both algorithms. \break

In the case of the singular peak, the agent commences at \textbf{\textit{initial=[0, 1]}} and for the multiple peaks, the agent initiates from position 1 \textbf{\textit{initial=[0.8, -0.5]}} and position 2 \textbf{\textit{initial=[0, -1]}}). We simulated a 3D depiction where the reward function aims to reach the peak, thus concentrating on a maximization problem. The agent is capable of moving (taking steps) across the grid-based environment such as North, South, East, West, as well as NE, NW, SE, SW, with a constant step size of $stepSize = [0.05, 0.05];$. This ensures consistent initial decision-making. 

\begin{equation}
o = -\sum_{i=1}^{n} x_i^2
\end{equation}

\begin{equation}
o = -1 \left(0.2 + x_1^2 + x_2^2 - 0.1 \cos(6\pi x_1) - 0.1 \cos(6\pi x_2)\right)
\end{equation}

For the SA algorithm, the optimal temperature and alpha values were determined to be 1 and 0.99, respectively, established through experimentation and tuning.

\begin{figure}[H]
\centering
\subfloat[\scriptsize{HC Implementation}]{
  \includegraphics[width=0.4\textwidth]{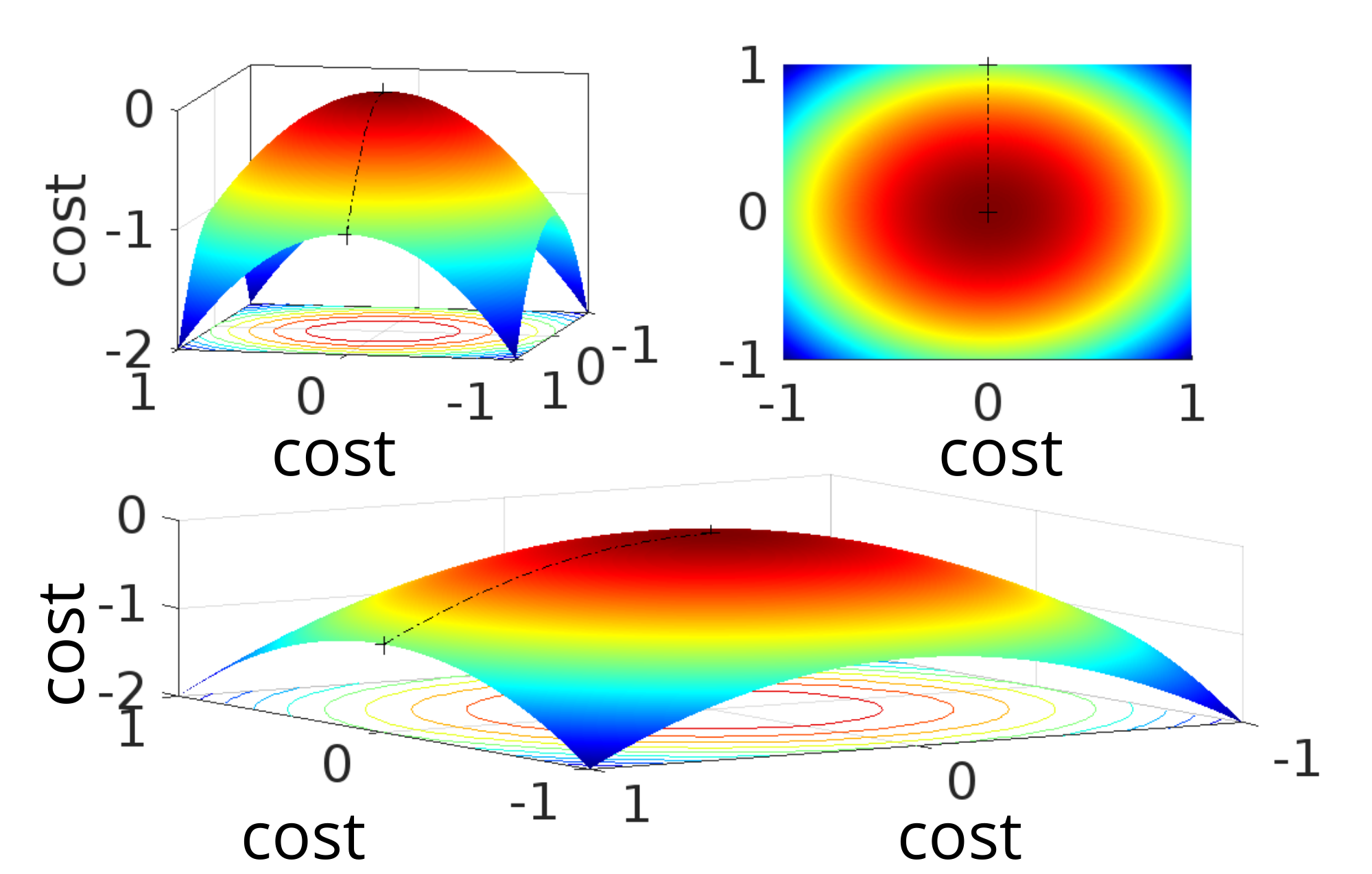}
  \label{fig:HCsubfig1}
}
\qquad
\subfloat[\scriptsize{SA Implementation}]{
  \includegraphics[width=0.4\textwidth]{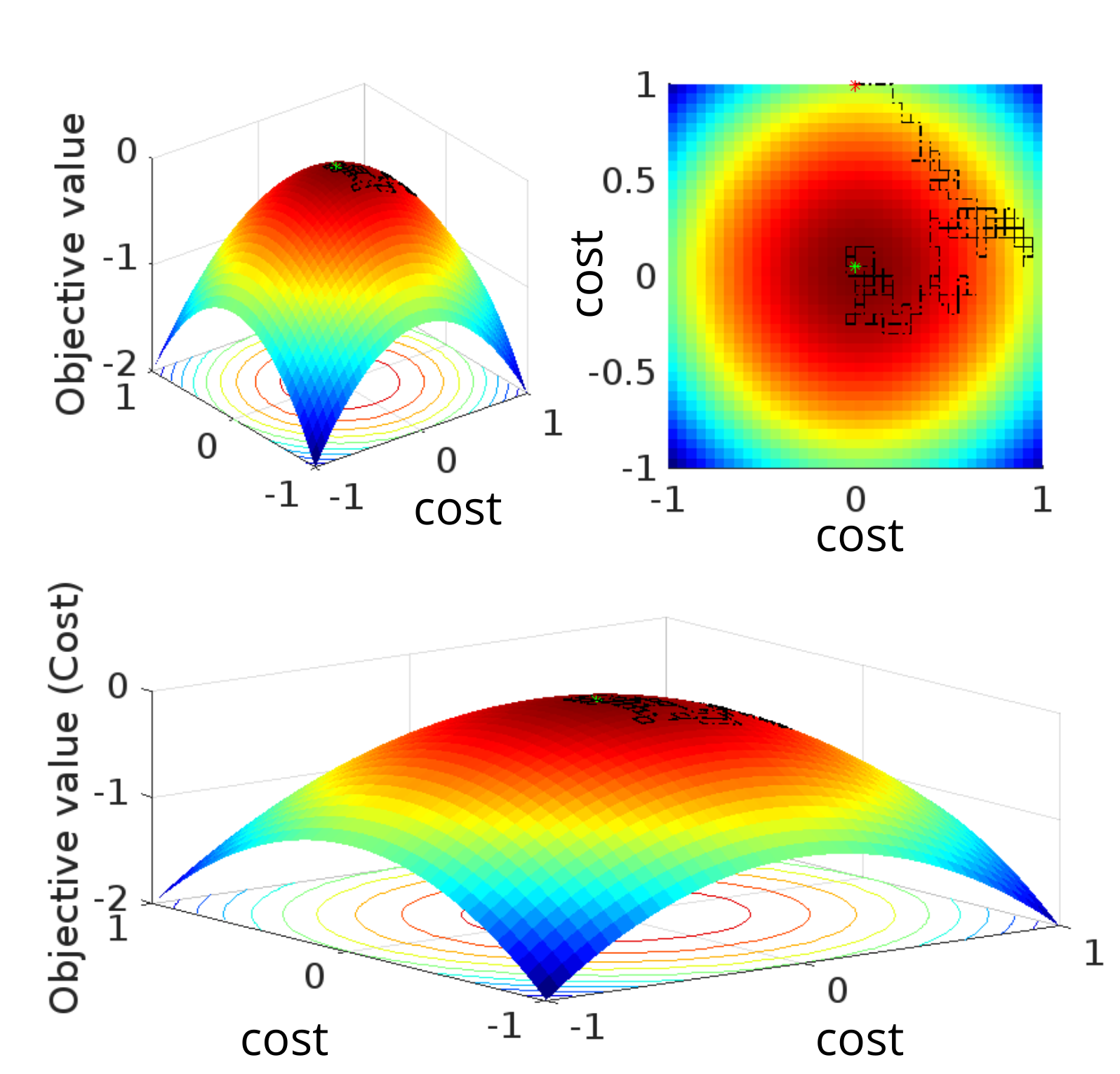}
  \label{fig:SAsubfig2}
}
\caption{Comparison between HC and SA by implementing Eq. 7. The axes are identical in all directions and represent a straightforward metric ranging from 1 to -1, indicating the length of the mesh.}
\label{fig:qwert}
\end{figure}

In Eq.~7's simulation, both methods reach a cost of 0, indicating they have achieved the peak (Fig.~\ref{fig:qwert}). 

The cost graph reveals that for a simple problem like this, HC is more suitable, as SA tends to overexploit the space, resulting in a less smooth cost vs iteration graph. This is also an initial insight into understanding the probability and temperature behaviors in SA implementation (Fig.~\ref{fig:cdcdcdc}):

\begin{figure}[H]
\centering
\subfloat[\scriptsize{HC cost graph}]{
  \includegraphics[width=0.4\textwidth]{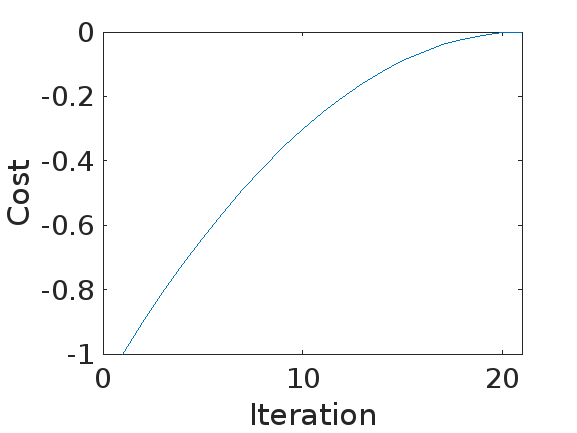}
  \label{fig:HCsubfig1_eq7}
}
\qquad
\subfloat[\scriptsize{SA temperature \& probability}]{
  \includegraphics[width=0.4\textwidth]{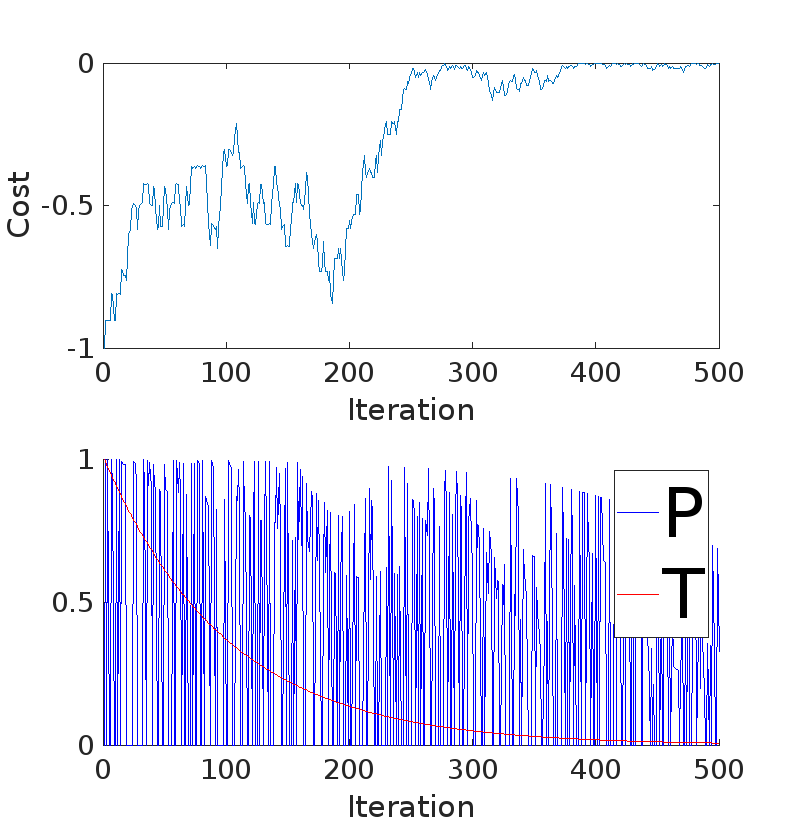}
  \label{fig:SAsubfig2_eq7}
}
\caption{Cost graph results for HC and SA using Eq. 7.}
\label{fig:cdcdcdc}
\end{figure}

In the process's early stages, when the temperature is high, the algorithm is more likely to accept worse solutions. This allows the algorithm to explore the search space and avoid getting stuck in local minima. As the temperature decreases, the agent approaches the goal (global minimum), and the probability of accepting worse solutions decreases. This steers the algorithm towards the best solution.\break

The temperature parameter in SA governs the equilibrium between exploration and exploitation. At the search's onset, the high temperature allows the algorithm to explore a vast range of solutions, both better and worse. As the search progresses and the agent approaches the goal, the temperature is gradually decreased. This 'cooling schedule' lessens the likelihood of accepting worse solutions and makes the search more concentrated. \break

\begin{figure}[H]
\centering
\subfloat[\scriptsize{HC implementation}]{
  \includegraphics[width=0.4\textwidth]{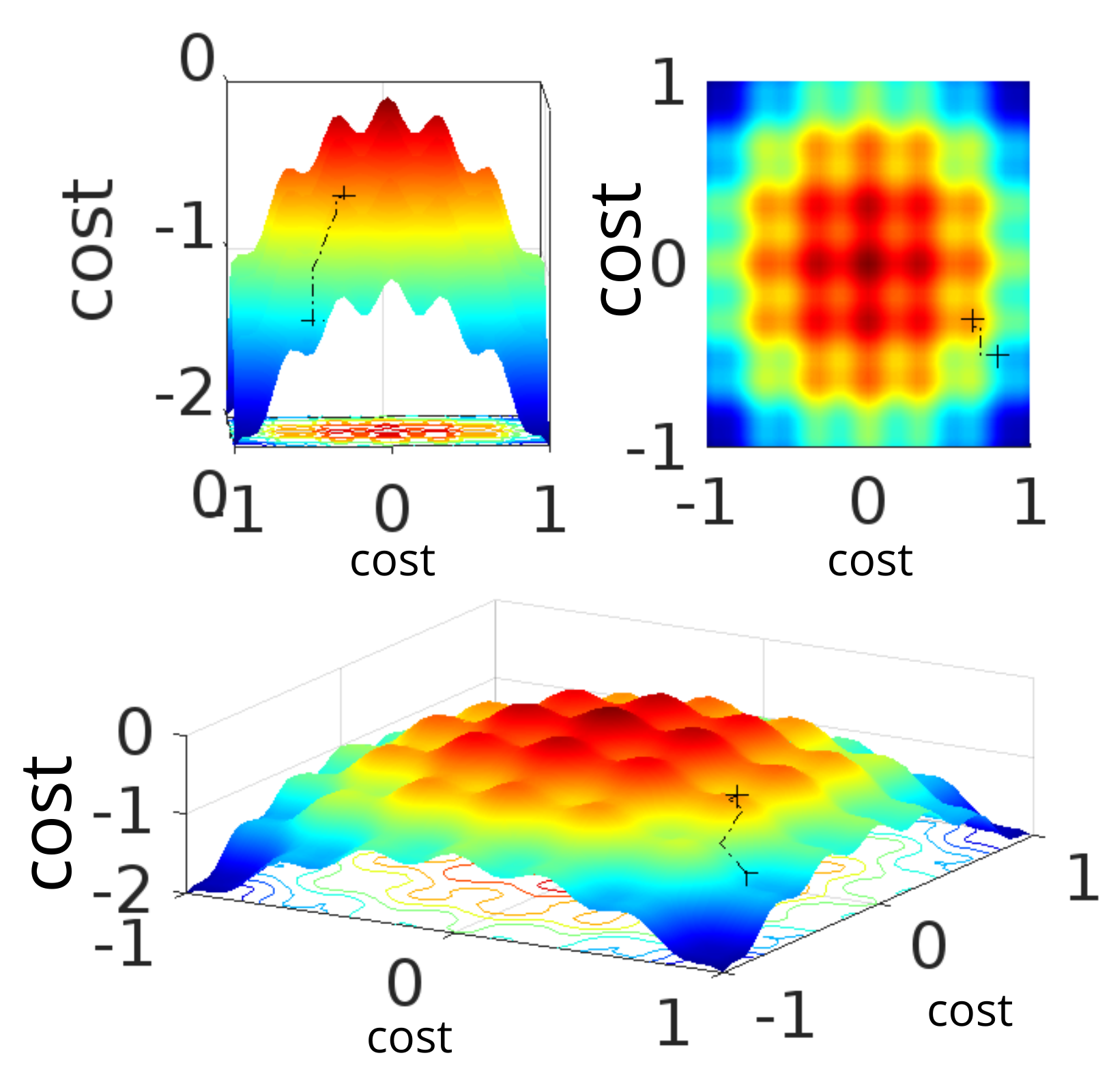}
  \label{fig:HCsubfig1_eq8}
}
\qquad
\subfloat[\scriptsize{SA implementation}]{
  \includegraphics[width=0.4\textwidth]{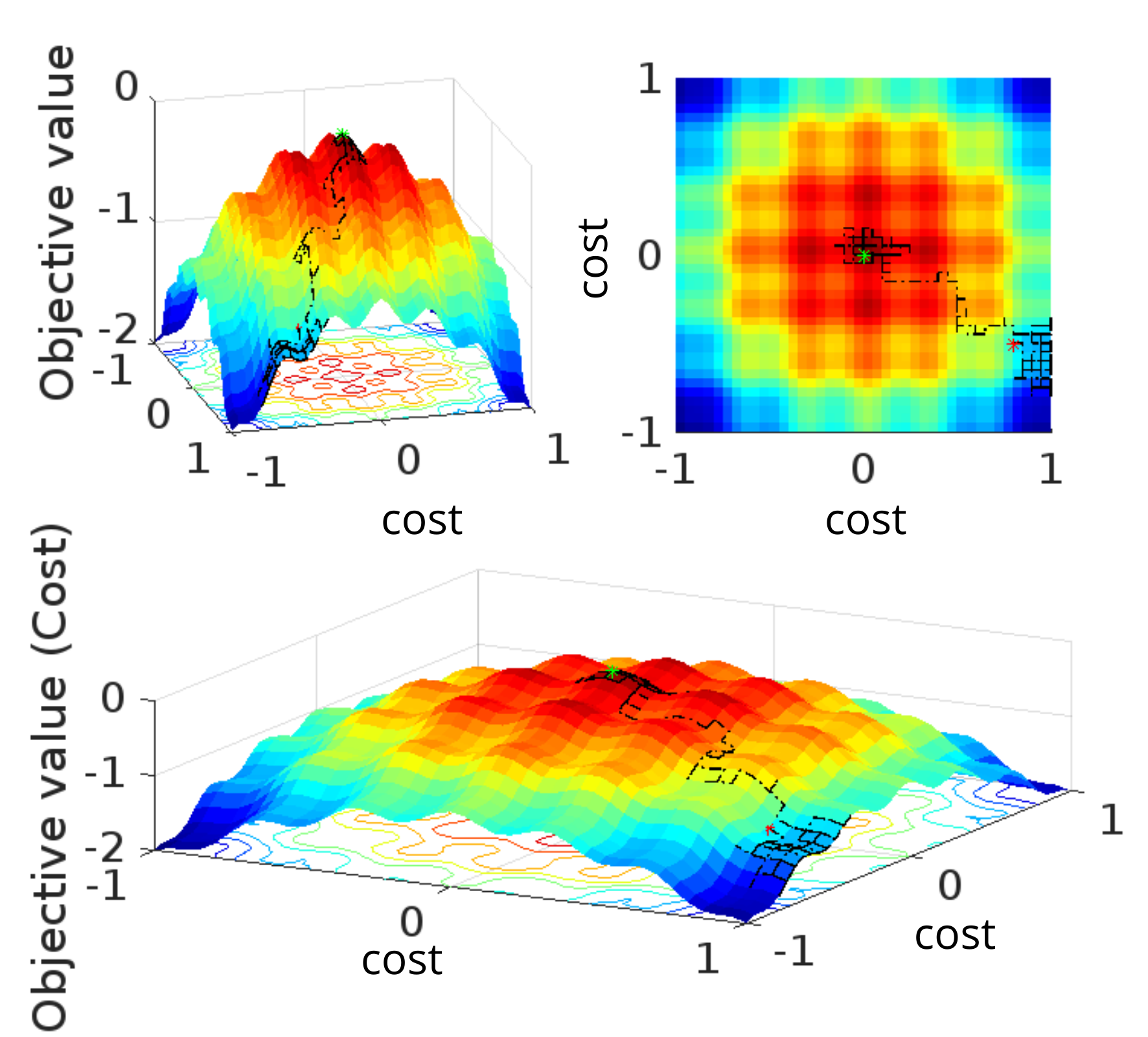}
  \label{fig:SAsubfig2_eq8}
}

\caption{Heat map comparison between HC and SA by Implementing Eq. 8 by starting on position 1. The axes are identical in all directions and represent a straightforward metric ranging from 1 to -1, indicating the length of the mesh.}
\label{fig:complex}
\end{figure}

\begin{figure}[H]
\centering
\subfloat[\scriptsize{HC cost graph}]{
  \includegraphics[width=0.4\textwidth]{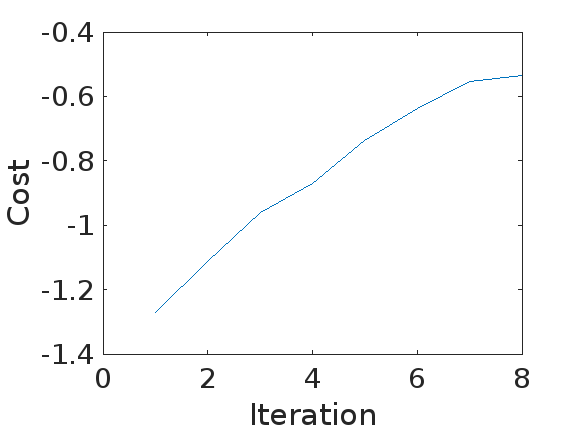}
  \label{fig:hcCOSTpos1}
}
\qquad
\subfloat[\scriptsize{SA temperature \& probability}]{
  \includegraphics[width=0.4\textwidth]{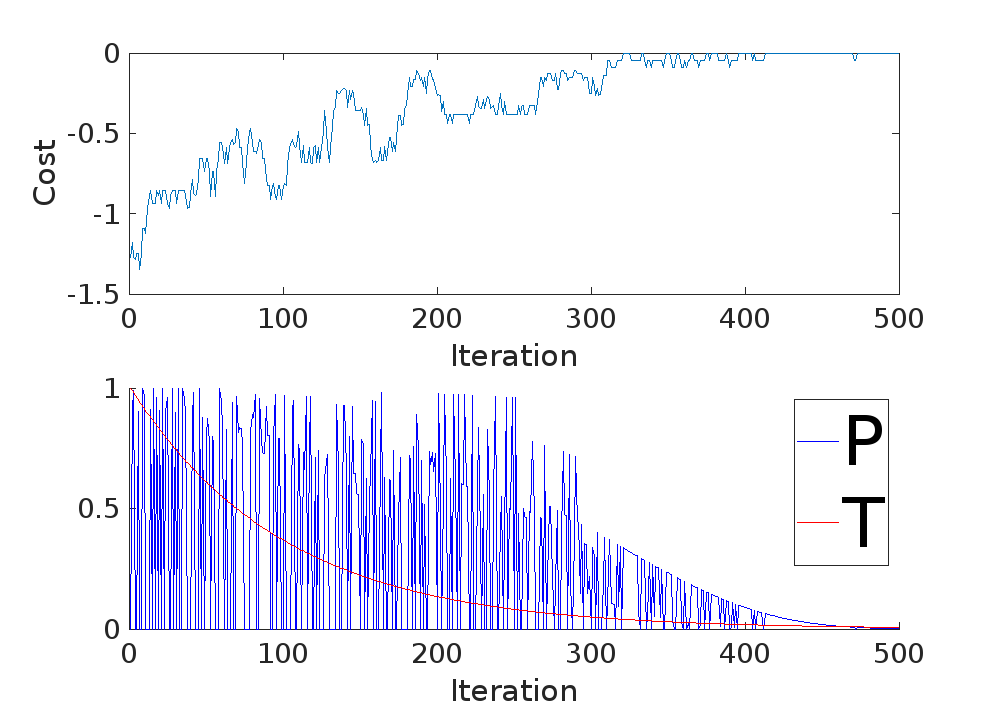}
  \label{fig:saCOSTpos1}
}

\caption{Cost graph results for HC and SA using Eq. 8 in position 1.}
\label{fig:figure}
\end{figure}

In scenarios where the search space incorporates multiple peaks, the HC agent faces challenges in identifying the global optimum, as depicted in Fig.~\ref{fig:HCsubfig1_eq8}. Conversely, the SA agent would be capable of pinpointing the global optimum. The following illustration (Eq. 8) showcases the agent's strategy of evading local optima to determine the most significant optimum within its region. Fig.~\ref{fig:SAsubfig2_eq8} delineate the navigational route the agent undertakes to ascend to the hill's apex. These figures also represent the evolution of its travel cost function alongside the diminishing temperature and probability management, as shown in Fig.~\ref{fig:saCOSTpos1}. The resulting graph clearly demonstrates the behavior described above, illustrating the impact of temperature and the exploration-exploitation trade-off in the simulated annealing algorithm and its attempts to break free from local optima, occasionally opting for inferior decisions even when better alternatives were previously available by accepting varying probabilities. It is critical to mention that achieving such a satisfactory solution required multiple simulation runs, leading to an optimal solution in approximately 1 out of 5 simulations. This behavior can be attributed to several factors, such as the Temperature Schedule decreasing too swiftly, thereby inhibiting the exploration of more regions. Additionally, the initial temperature may be too low, leading to a premature convergence. Other contributing factors include multiple simulation runs and the complexity of the Neighborhood Function. It is also observed that the agent tends to favor horizontal (North, South, East, West) rather than vertical movements, possibly to avoid the increased complexity of the chosen route (Fig.~\ref{fig:complex}). \break

Hill Climbing starts from an arbitrary solution and iteratively makes small changes to improve it. However, it often gets stuck in local optima because it lacks a mechanism to explore other regions of the solution space once it reaches a peak. As a result, the algorithm may exhibit significant variations as it explores different solutions, leading to a more erratic cost curve. The improvement is more gradual and less erratic when a higher number of sampling points is considered. \break

On the other hand, Simulated Annealing utilizes a temperature-controlled mechanism, which allows it to escape local optima and increase the likelihood of finding the global optimum. This is achieved by occasionally accepting ”bad” moves (moves that lead to worse solutions) to explore more of the solution space. As the ”temperature” decreases over time, the algorithm becomes less likely to accept bad moves and more inclined to exploit the regions of the solution space that have been identified as good, with the hope of converging on the global optimum.

\section{Experimental Results and Validation}

\subsection{Experiment Set-up}

\subsubsection{Hardware Set-up }
\begin{figure}[H]
\centering
\includegraphics[width=\columnwidth, height=0.5\columnwidth, keepaspectratio]{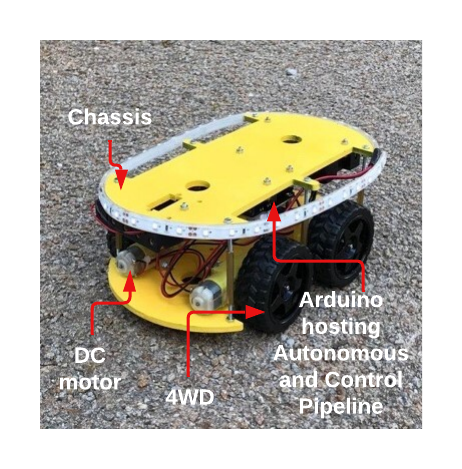}
\caption{Miniature size of ROMIE for testing the pipeline and mechanical properties.}
\label{fig:romiemini}
\end{figure}

The GPP testing and implementation is conducted using a Jetson Orion development kit and a
Lenovo ThinkPad P53 equipped with an Intel Core i7-9750H CPU running at 2.6GHz (12 cores) and
an Nvidia Corporation TU177GLM GPU. The ThinkPad P53 also has 16GiB of memory. The prototyped ROMIE is presented in Fig.~\ref{fig:romiemini}.

\subsubsection{Software language}

Python and MATLAB are effective language for solving the Traveling Salesman Problem (TSP) due to their advanced numerical abilities, user-friendly syntax, and high-level problem-solving abstractions. Both have valuable features: Python includes packages like Numpy and Panda for array management, and MATLAB has a robust mathematical toolbox. \break

While JavaScript is used for web application backend functionalities, compiled languages like Java or C++ might be chosen when faster execution is needed. However, these languages typically require more coding. \break

Despite the potential speed advantages of Java and C++, the report primarily focuses on Python and MATLAB because of their popularity among engineers and the abundance of open-source TSP models they offer. \break

Python and MATLAB are considered effective languages for solving the TSP because of their robust numerical and computational capabilities. Both languages provide user-friendly syntax and high-level abstractions which make them ideal for algorithmic problem solving. Also, JavaScript is employed to implement various backend functionalities
within our web application.
Python is a high-level language that offers various packages, including Numpy and Panda, which
are valuable for array management. Similarly, MATLAB provides a robust mathematical toolbox.
However, when performance is a critical factor, languages like Java or C++ can also be employed
for TSP problem solving. These compiled languages often yield faster execution times compared
to interpreted languages like Python and MATLAB. Nonetheless, they require more coding implementation. This report primarily focuses on Python and MATLAB models, as they are favored by
engineers and offer numerous open-source TSP models.

    \subsubsection{Webframe work}
Django was chosen as the web framework for the project due to its numerous advantages linked with Python capabilities. Its major qualities include its robustness, scalability, and extensive
built-in features that facilitate rapid development.

    \subsubsection{Data set}
    
The report utilizes six distinct datasets, featuring 20, 100, 200, and 1000 sampling points respectively (some graphs will show Tabu Search (TS) which was already computed).  Each dataset comprises longitude and latitude coordinates, illustrating the region of Endeavour Mining's Ity Mine, located in Ivory Coast. The selection of this specific location facilitates practical application, enabling ROMIE to work with real datasets that mining engineers and geologists might select to carry out their prospecting efforts.

\subsection{Google OR-Tools VS RL comparative table}

The results have been computed for three different methods: OR-Tools, Q-Learning and Double Q-Learning. Tables \ref{tabel_results_1} and \ref{tabel_results_2} provide a summary of the results obtained during the simulation by running each algorithm a subsequent amount of time (between 10 and 100 times) and taking the mean of the final runs. The
analysis of the results is provided in the subsequent sections to evaluate the performance of TSP model. All bolded values in this table represent the best result for each instance of different TSP experiments. One key aspect to understand from this table is the percentage gap, which represents the difference between the best value obtained in a TSP run and the results of all other algorithms that performed worse. This gap indicates how far each algorithm's performance deviates from the best result and how closely the results of different algorithm implementations compare to each other. Each TSP instance used the same set of waypoint coordinates to ensure a fair and consistent comparison, with every run starting from the same coordinates. The final coordinate is not predetermined, allowing the algorithm complete freedom to choose its endpoint.

\begin{table}[H]
\centering
\caption{Overall TSP results for running the pre-defined dataset using Google OR-Tools and Q-Learning and Double Q-Learning on a prospecting land of 11km$^2$. The Tour Len is represented in meters.}
\label{tabel_results_1}
\begin{tabular}{|l| l p{0.5in} l |l p{0.5in} l |l p{0.5in} l|}
\hline
& \multicolumn{3}{c}{TSP20} & \multicolumn{3}{c}{TSP50} & \multicolumn{3}{c}{TSP100} \\
\hline
Method & Tour Len. & Gap to best (\%) & Time (s) & Tour Len. & Gap to best (\%) & Time (s) & Tour Len. & Gap to best (\%) & Time (s) \\
\hline

PCA & 12951 & 16.72 & 0.003 & 18440.19 & 0.44 & 0.026 & 26093.88 & 6.1 & 0.18 \\
\hline

PMCA & 12951 & 16.72 & 0.004 & 18440.19 & 0.44 & 0.022 & 26093.88 & 6.1 & 0.18 \\

\hline

LCI & 12949.2 & 16.7 & 0.005 & 18654.64 & 1.6 & 0.020 & 26132.7 & 6.3 & 0.09 \\

\hline
GCA & 12951.3 & 16.72 & 0.005 & 19241.21 & 4.8 & 0.051 & 26735.3 & 8.7 & 0.16 \\

\hline
LCA & 12951.3 & 16.72 & 0.004 & 18548.36 & 1.0 & 0.037 & 25931.26 & 5.5 & 0.16 \\

\hline
FUMV & 12951.3 & 16.72 & 0.006 & 18884.45 & 2.86 & 0.047 & 27911.1 & 13.5 & 0.22 \\

\hline
S & 12951.3 & 16.72 & 0.004 & 18872.06 & 2.79 & 0.035 & 26853.75 & 9.2 & 0.12 \\

\hline
C & 12951.3 & 16.72 & 0.003 & 19769.34 & 7.67 & 0.028& 26211.44 & 6.6 & 0.18 \\

\hline
PCI & 12951.3 & 16.72 & 0.005 & 18862.34 & 2.73 & 0.054& 27351.69 & 11.2 & 0.15  \\


\hline
GD & 13306 & 19.91 & 0.003& 18978 & 3.36 & 0.012& 26233 & 6.7 & 0.118  \\

\hline
GLS & 13306 & 19.91 & 0.002& 18650 & 1.56 & 0.9& 25662 & 4.4 & 33.4 \\

\hline
SA & 13306 & 19.91 & 0.002& 18978 & 3.36 & 0.01504 & 26088 & 6.1 & 0.12\\

\hline
TS & 13306 & 19.91 & 0.002& 18652 & 1.59 & 0.0584& 26088 & 6.1 & 0.0115 \\

\hline
GTS & 13306 & 19.91 & 0.002& 18978 & 3.36 & 0.015& 26088 & 6.1 & 0.125\\

\hline
Q-L & \textbf{11096.2} & \textbf{-} & \textbf{0.02} & \textbf{18360.3} & \textbf{-} & \textbf{0.07} & \textbf{24588.1} & \textbf{-} & \textbf{0.2456}  \\

\hline

D Q-L &  11096.2 &  0.05& 0.03& 18360.3&  - &  0.09&  25367.87&  4.5&  0.4 \\

\hline
\end{tabular}
\end{table}

\begin{table}[H]
\centering
\caption{Overall TSP results for running the pre-defined dataset using Google OR-Tools and Q-Learning and Double Q-Learning on a prospecting land of 11km$^2$. The Tour Len is represented in meters.}
\label{tabel_results_2}
\begin{tabular}{|l| l p{0.5in} l| l p{0.5in} l |l p{0.5in} l|}
\hline
& \multicolumn{3}{c}{TSP200} & \multicolumn{3}{c}{TSP500} & \multicolumn{3}{c}{TSP1000} \\
\cline{1-4} \cline{5-7} \cline{8-10}
Method & Tour Len. & Gap to best (\%) & Time (s) & Tour Len. & Gap to best (\%) & Time (s) & Tour Len. & Gap to best (\%) & Time (s) \\
\hline

PCA & 38560.46 & 7.2 & 0.807 & 57468.88 & 0.70 & 7.2& 80708.54 & 1.38 & 55.89(  \\
\hline

PMCA  & 38048.31 & 5.8 & 0.882& 57468.88 & 0.70 & 7.25 & 80404.24 & 1.0 & 59.87 \\
\hline

LCI & 38345.30 & 6.6 & 0.469& 59746.03 & 4.69 & 11.01& 82143.63 & 3.19 & 48.48 \\
\hline

GCA  & 37380.11 & 3.9 & 1.26& 58262.26 & 2.09 & 10.53& 81282.79 & 2.11 & 62.79 \\
\hline

LCA & 38531.86 & 7.1 & 0.93& 58902.07 & 3.21 & 11.92& 81873.33 & 2.855 & 52.41 \\
\hline

FUMV & 38467.23 & 7.0 & 1.41& 58133.75 & 1.86 & 11.22& 81600.11 & 2.50 & 70.69 \\
\hline

S & 39210.86 & 9.0 & 0.66& 57468.88 & 0.70 & 7.04& 83508.45 & 4.90 & 36.09 \\
\hline

C  & 37739.44 & 4.9 & 0.68& 57570.74 & 0.88 & 10.89& 79703.71 & 0.12 & 49.42 \\
\hline

PCI & 39596.75 & 10.1 & 0.74 & 59642.32 & 4.51 & 7.78& 82426.53 & 3.54 & 44.76 \\
\hline

GD  & 38486 & 7.0 & 0.595& 57127 & 0.10 & 5.97 & 80003 & 0.49 & 50 \\
\hline

GLS & 36861.5 & 2.5 & 108 & \textbf{57068} & \textbf{-} & \textbf{10} & 79919 & 0.39 & 92 \\
\hline

SA & 38486 & 7.0 & 0.4& 57872 & 1.41 & 4.94 & 79607 & - & 52.5 \\
\hline

TS & 38247 & 6.4 & 19.919 & 57872 & 1.41 & 5.05& 79607 & - & 49.893 \\
\hline

GTS & 38486 & 7.0 & 0.594& 57872 & 1.41 & 5.33& \textbf{79607} & \textbf{-} & \textbf{49.745}  \\
\hline

Q-L & \textbf{35962.36} & \textbf{-} & \textbf{13.2} & 57830 & 1.36 & 290 & 84589 & 6.26 & 600  \\
\hline

D Q-L &  35985.32 &  0.1&  16.3&  59830.26&  3&  106&  84647&  6&  660\\

\hline
\end{tabular}
\end{table}

Table~\ref{tabel_results_std} displays statistical data for the two RL algorithms used in this study: QL and DQL.

Due to the inherent randomness in the Q-table initialization process, each run of the algorithm can produce varying results. Unlike Google OR-Tools' heuristic approach, which does not involve random initialization, RL algorithms are unlikely to yield identical results in repeated runs. As the Q-table becomes more populated with values, the algorithm's performance tends to degrade. Therefore, it's crucial to evaluate statistical measures like the standard deviation, which quantifies the variation or dispersion within a set of values. A lower standard deviation indicates values closer to the mean (average), suggesting better robustness. By analyzing these statistical features, we gain insights into the algorithm's robustness and the cost implications of tour length variations. This information is valuable for assessing the engineering and economic impact of these variations on the ROMIE project. \break

\begin{table}[h!]
\centering
\caption{Reinforcement Learning Standard Deviation (std) and Variance (var) Results}
\label{tabel_results_std}
\begin{tabular}{|c|c|c|c|c|c|c|c|c|c|c|c|c|c|}
\hline
\textbf{Algorithm} & \textbf{Method} & \multicolumn{2}{c|}{\textbf{TSP20}} & \multicolumn{2}{c|}{\textbf{TSP50}} & \multicolumn{2}{c|}{\textbf{TSP100}} & \multicolumn{2}{c|}{\textbf{TSP200}} & \multicolumn{2}{c|}{\textbf{TSP500}} & \multicolumn{2}{c|}{\textbf{TSP1000}} \\ \hline
 & & \textbf{TL} & \textbf{T} & \textbf{TL} & \textbf{T} & \textbf{TL} & \textbf{T} & \textbf{TL} & \textbf{T} & \textbf{TL} & \textbf{T} & \textbf{TL} & \textbf{T} \\ \hline
\textbf{std} & QL & 49 & 0.2 & 132 & 0.4 & 216 & 0.4 & 149 & 0.6 & 335 & 2 & 669 & 3 \\ \cline{2-14} 
 & DQL & 32 & 0.2 & 178 & 0.3 & 212 & 0.4 & 161 & 0.9 & 720 & 0.8 & 776 & 1.6 \\ \hline
\textbf{var} & QL & 2356 & 0.1 & 17304 & 0.2 & 46510 & 0.1 & 22140 & 0.4 & 112404 & 4.2 & 447473 & 8.7 \\ \cline{2-14} 
 & DQL & 991 & 0.1 & 31673 & 0.1 & 79329 & 0.2 & 26060 & 0.8 & 518986 & 0.6 & 602475 & 2.4 \\ \hline
\end{tabular}
\label{tab:tsp_comparison}
\end{table}

\subsection{Google OR-Tools Implementation and Results}

Google OR-Tools algorithms are partitioned into two categories: First Solution (FS) Strategies and Local Search (LS) Meta-heuristics. Broadly speaking, the performance of first solution strategies aligns closely with certain Local Search Metaheuristics. Table~\ref{tab:results} provides a comparative evaluation that we generated of Google OR-Tools' performance in terms of first solution and local search strategies relative to the optimal performance from the dataset. We assess the optimal performance of a dataset by establishing the best observed results as the benchmark and calculating the Gap to Best (GB).

\begin{table}[H]
\centering
\caption{Average Google OR-Tools Results Across The Dataset For Each Solution Strategy}
\label{tab:results}
\begin{tabular}{|l| l l l l l l|}
\hline
& TSP20 & TSP50 & TSP100 & TSP200 & TSP500 & TSP1000 \\
\hline
FS & 16.7\% & 2.7\% & 8.1\% & 6.9\% & 2.2\% & 2.4\% \\
LS & 19.9\% & 2.7\% & 5.9\% & 6.0\% & 1.1\% & 0.2\% \\
\hline
\end{tabular}
\end{table}

\begin{figure}[H]
\centering
\includegraphics[width=1\columnwidth, keepaspectratio]{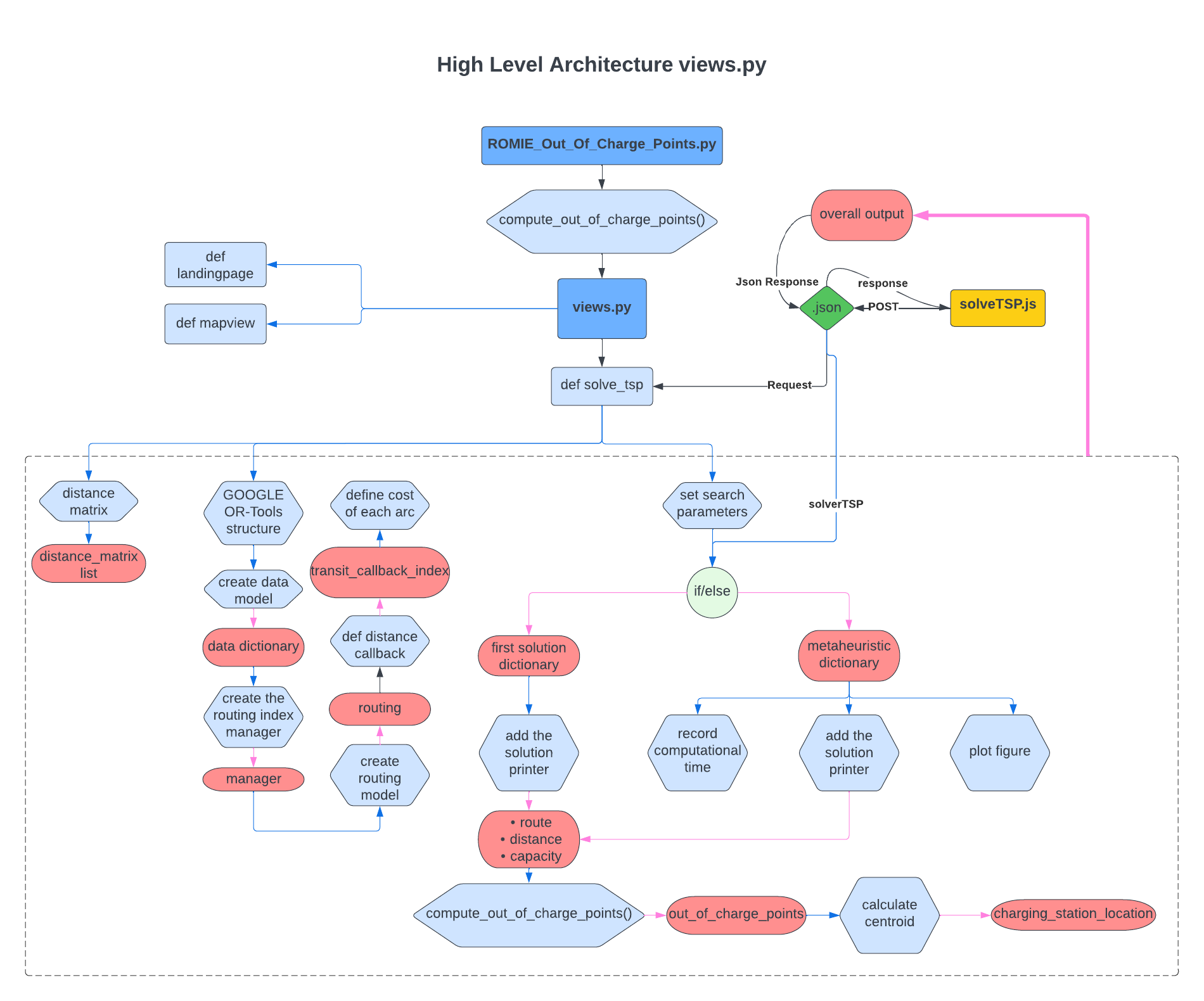}
\caption{Google OR-Tools implementation}
\label{ORTOOO}
\end{figure}

The performance of each meta-heuristic set can be visualized through a graph depicting the amount of time required to compute the best TSP tour by taking the running average to obtain smoother results. The red circle shows the cost effective distance results for each methods and the blue circle shows the overall cost effective distance result of the data set. The implementation within the code can also be visualised via Fig.~\ref{ORTOOO}.

\begin{figure}[H]
    \centering
        \subfloat[\scriptsize{Google OR-Tools for TSP20}]{
          \includegraphics[width=0.4\textwidth]{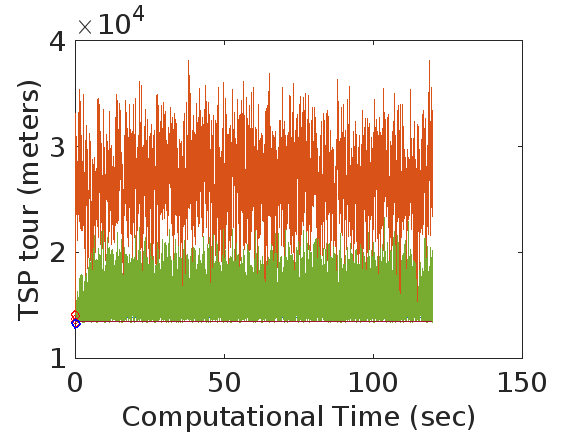}
          \label{fig:ortsp20}
        }
        \qquad
        \subfloat[\scriptsize{Google OR-Tools for TSP20 ZOOM}]{
          \includegraphics[width=0.4\textwidth]{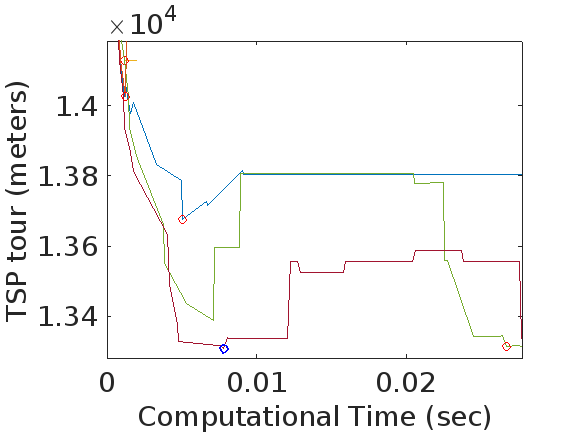}
          \label{fig:ortsp20ZOOM}
        }
        \qquad
        \subfloat[Google OR-Tools for TSP50]{
          \includegraphics[width=0.4\textwidth]{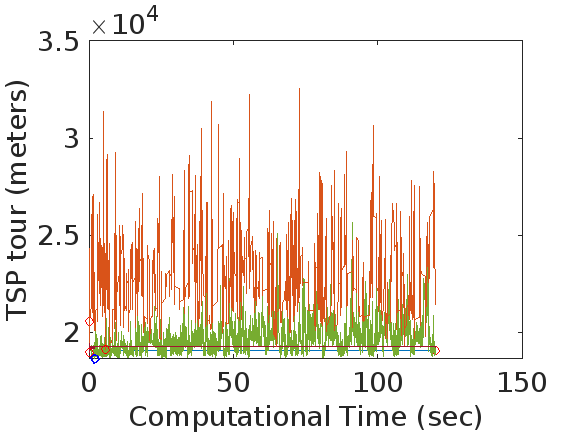}
          \label{fig:ortsp50}
        }
        \qquad
        \subfloat[Google OR-Tools for TSP50 ZOOM]{
          \includegraphics[width=0.4\textwidth]{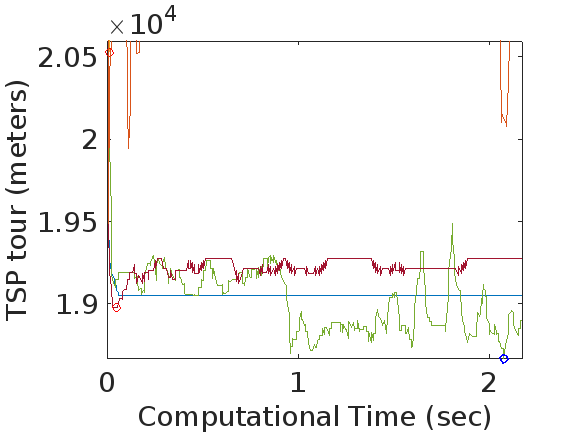}
          \label{fig:ortsp50ZOOM}
        }
        \qquad
        \subfloat[\scriptsize{Google OR-Tools for TSP100}]{
          \includegraphics[width=0.4\textwidth]{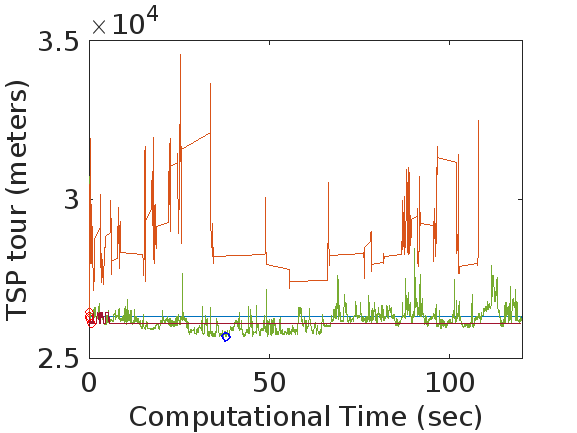}
          \label{fig:ortsp100}
        }
        \qquad
        \subfloat[\scriptsize{Google OR-Tools for TSP100 ZOOM}]{
          \includegraphics[width=0.4\textwidth]{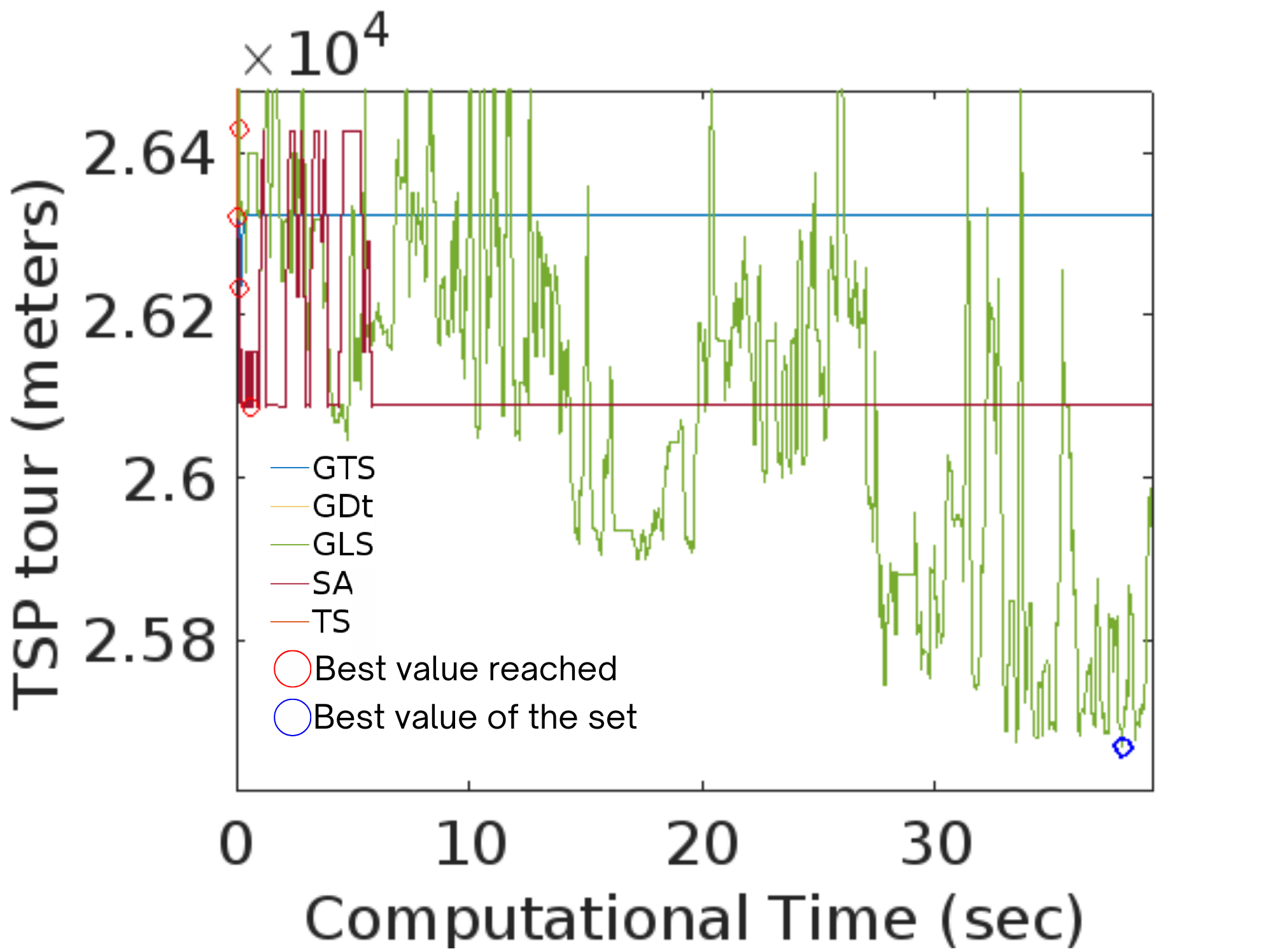}
          \label{fig:ortsp100ZOOM}
        }
    \caption{Cost of the TSP tour versus computational time using Google OR-Tools method.}
    \label{fig:enter-label}
\end{figure}

\begin{figure}[H]
    \centering
        \subfloat[\scriptsize{Google OR-Tools for TSP200}]{
          \includegraphics[width=0.4\textwidth]{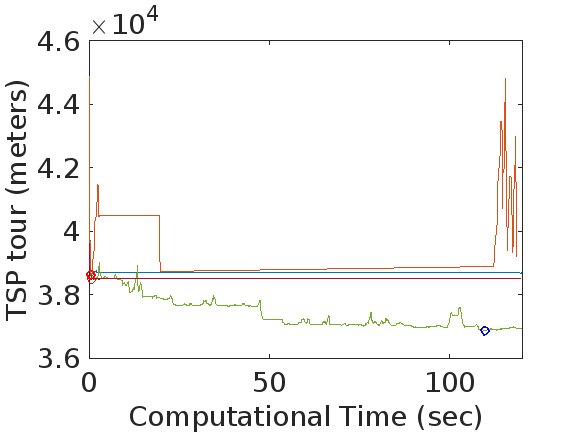}
          \label{fig:ortsp200}
        }
        \qquad
        \subfloat[\scriptsize{Google OR-Tools for TSP200 ZOOM}]{
          \includegraphics[width=0.4\textwidth]{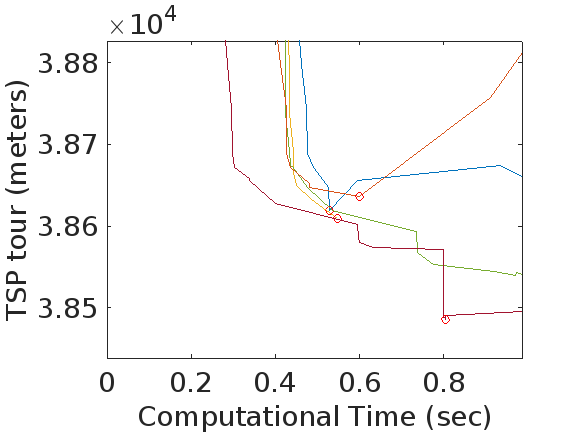}
          \label{fig:ortsp200ZOOM}
        }
        \qquad
        \subfloat[Google OR-Tools for TSP500]{
          \includegraphics[width=0.4\textwidth]{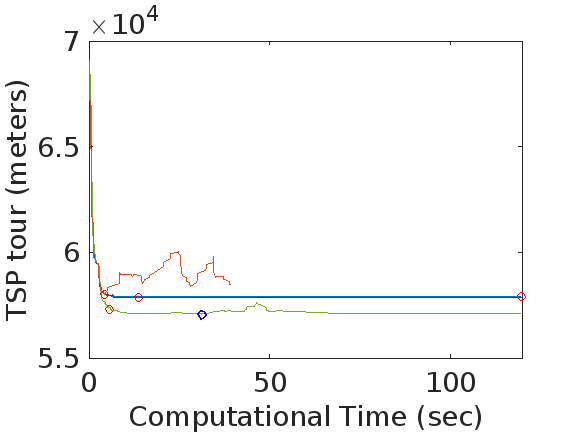}
          \label{fig:ortsp500}
        }
        \qquad
        \subfloat[Google OR-Tools for TSP500 ZOOM]{
          \includegraphics[width=0.4\textwidth]{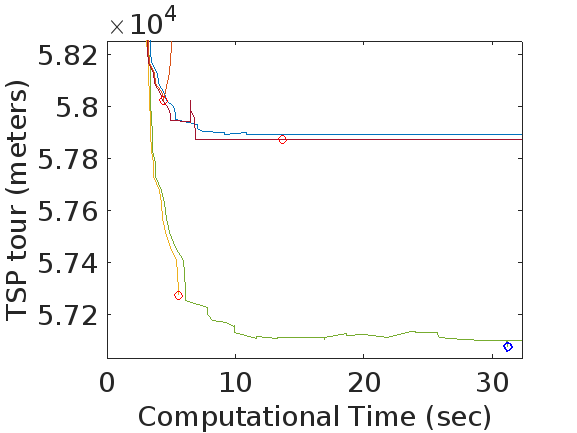}
          \label{fig:ortsp500ZOOM}
        }
        \qquad
        \subfloat[\scriptsize{Google OR-Tools for TSP1000}]{
          \includegraphics[width=0.4\textwidth]{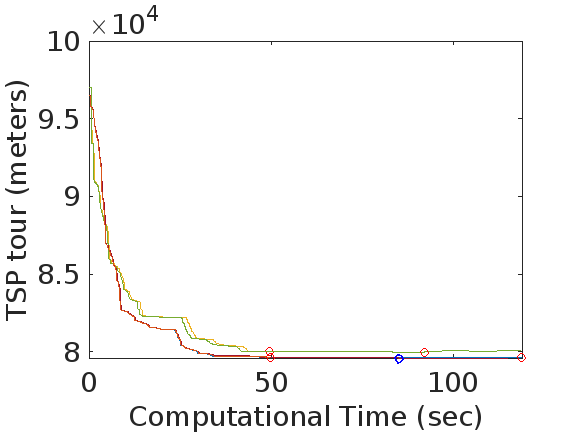}
          \label{fig:ortsp1000}
        }
        \qquad
        \subfloat[\scriptsize{Google OR-Tools for TSP1000 ZOOM}]{
          \includegraphics[width=0.4\textwidth]{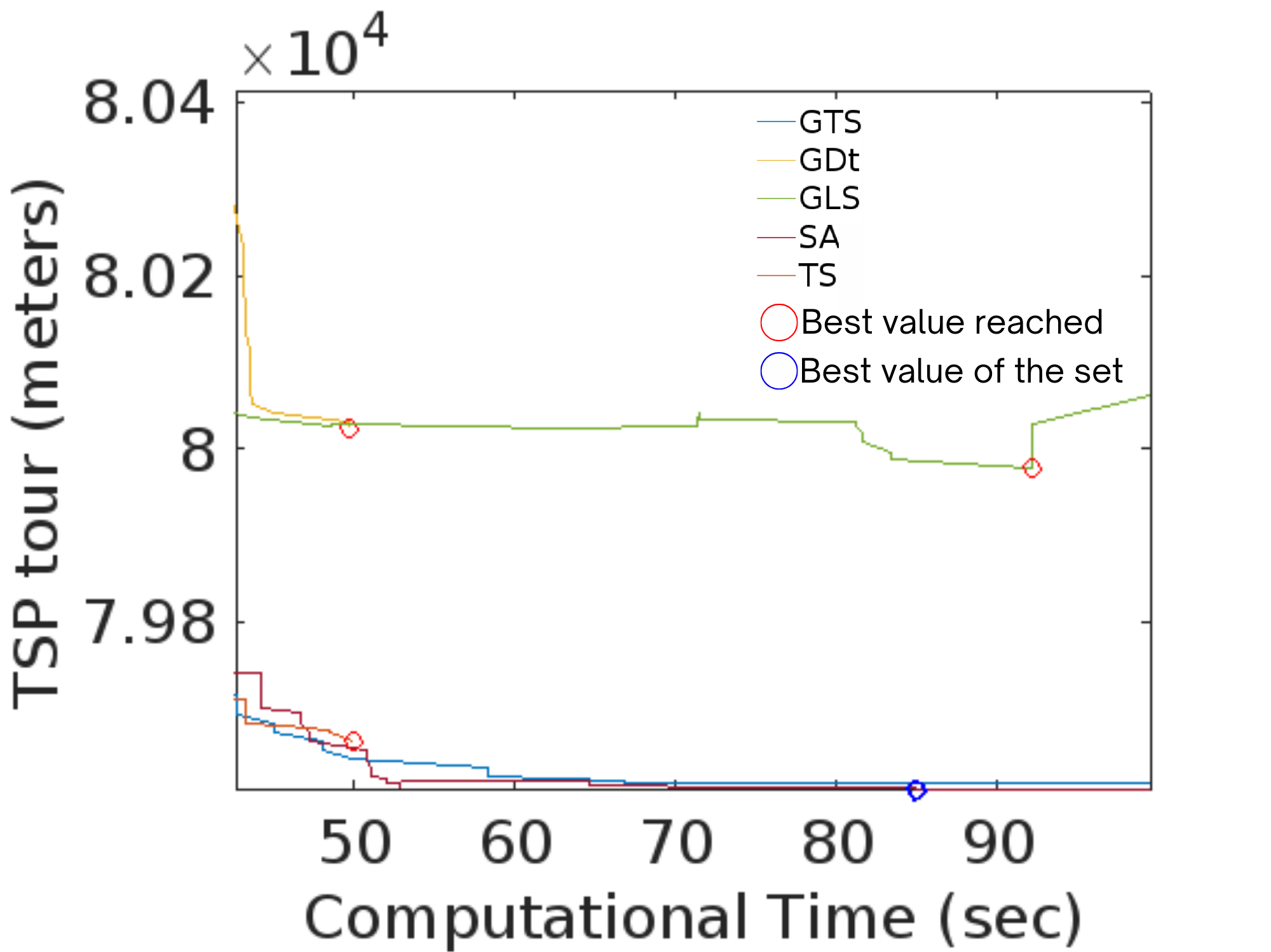}
          \label{fig:ortsp1000ZOOM}
        }
    \caption{Cost of the TSP tour versus computational time using Google OR-Tools method.}
    \label{fig:ZOOM}
\end{figure}

These charts elucidate the stability and variability of each method for a specified quantity of sampling points aimed at determining the optimal TSP tour. As verified earlier in our comparison between Simulated Annealing (SA) and Hill Climbing (HC), it is projected that with an increased count of sampling points, the agent tends to break free from local optima more swiftly than HC, even if the disparity is marginal for larger data sets (0.3\%). Nonetheless, the promptness of a method's convergence to the objective function underscores its stability and precision, thus ensuring a reliable forecast for subsequent computations of the best TSP tour. For a smaller set of sampling points, the meta-heuristic method seems to over-complicate the problem, resulting in a highly variable pattern as it gets trapped in local optima.

The Gradient Descent (GD) method typically halts its computation first, often finding itself ensnared in local optima (refer to Fig.~\ref{fig:ZOOM} for the yellow line).

\begin{enumerate}
\item \textbf{TSP20}: Results demonstrate a highly inconsistent and fluctuating Generalized Local Search (GLS) and Tabu Search (TS), in contrast to Guided Tabu Search (GTS) and SA, which rapidly locate the optimal solution and evade local optima (Fig.~\ref{fig:ortsp20}).
\item \textbf{TSP50}: Similar observations can be made for GLS, TS, SA, and GTS. Despite its erratic behavior, GLS appears to find the optimal solution in this iteration(Fig.~\ref{fig:ortsp50}).
\item \textbf{TSP100}: As we increase the number of sampling points, we see marginal improvements in the behavior of GLS and TS. GTS quickly avoids local optima, but SA seems to have more difficulty escaping local optima, exhibiting slight volatility in the initial 5 seconds. SA and GLS showcase a more binary behavior in the zoomed version (Fig.~\ref{fig:ortsp100ZOOM}). GLS achieves the best solution for this round.
\item \textbf{TSP200}: A clear differentiation is seen here with SA, TS, and GTS struggling to avoid local optima, and an unusual exploration-exploitation tradeoff in TS. Nonetheless, GLS consistently seems to escape local optima as the problem's complexity (i.e., more sampling points) increases, resulting in the best outcome within this dataset.
\item \textbf{TSP500}: SA and GTS display similar outcomes with comparable behaviors. Surprisingly, GD proves to be more efficient than these two methods in this instance, potentially due to fortuitous exploration and exploitation. However, this may not always be the case given the method's instability. GLS also dominates this round, showcasing intriguing behavior around the 45-second mark when it explores other paths before returning to its optimal one (Fig.~\ref{fig:ortsp500}).
\item \textbf{TSP1000}: This is the sole instance where all methods exhibit smooth behavior, demonstrating precision in terms of exploration and exploitation. However, GTS outperforms the others, closely followed by the remaining methods. GTS's success can be attributed to its strategy of maintaining a "tabu list" of recent solutions and forbidding or penalizing moves that revert to these solutions. This approach is often more effective at finding global optima in complex solution spaces, which may explain its superior performance here (Fig.~\ref{fig:ortsp1000}).
\end{enumerate}

\begin{figure}[H]
    \centering
    \includegraphics[width=0.5\textwidth]{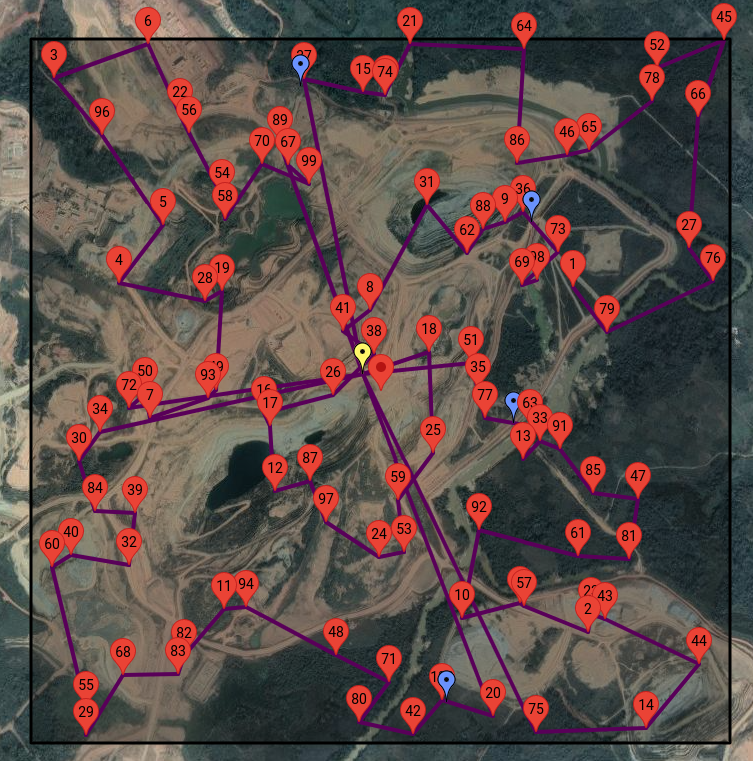}
    \caption{Google OR-Tools TSP100.}
    \label{fig:ORtsp100}
\end{figure}

Fig.~\ref{fig:ORtsp100} showcases the results of TSP 100 computed by Google OR-Tools.

\subsection{RL Implementation and Results}

\begin{figure}[H]
    \centering
    \includegraphics[width=1\textwidth]{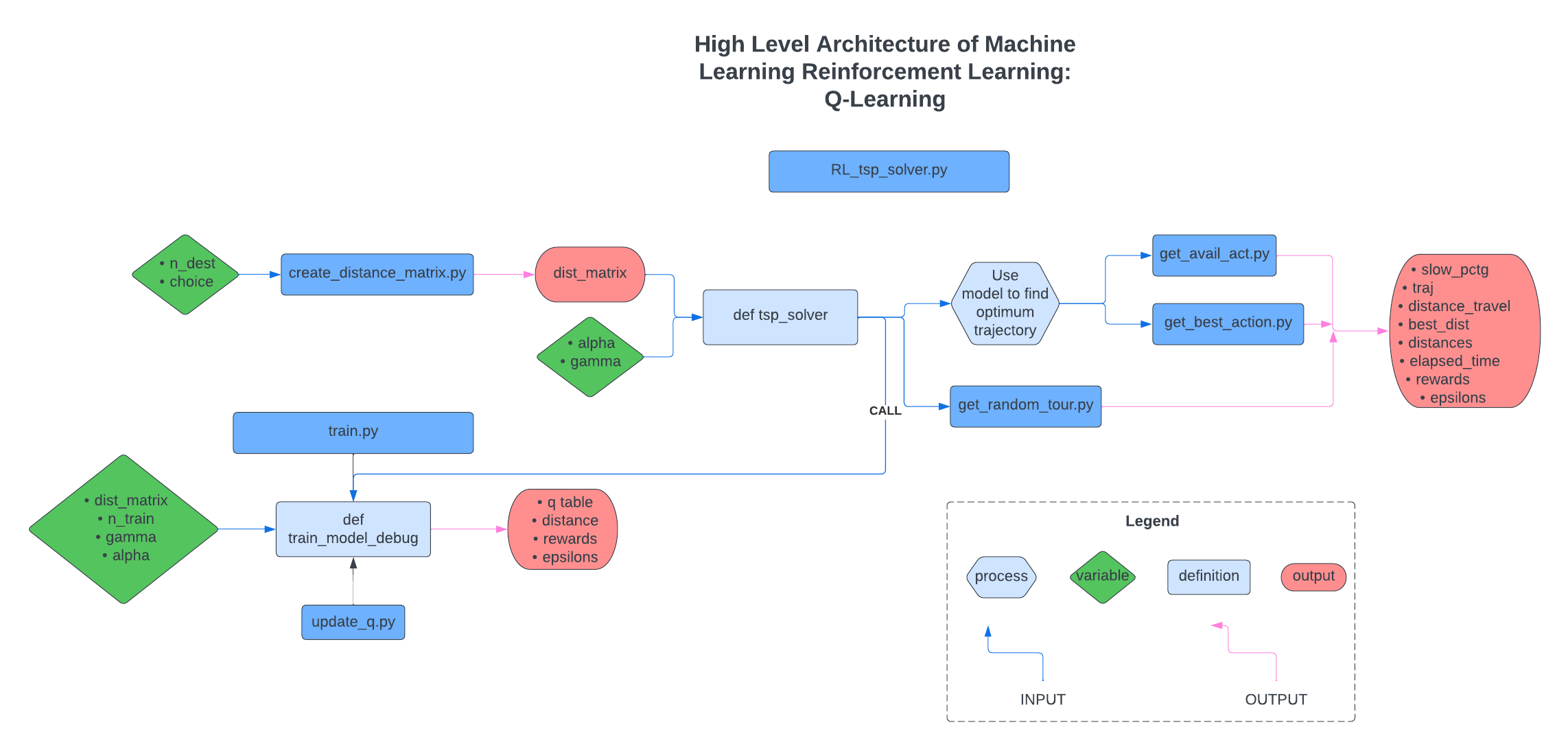}
    \caption{Reinforcement Learning Implementation}
    \label{RL}
\end{figure}

The results from both Q-Learning and Double Q-Learning are highly pleasing, achieved through diligent hyperparameter adjustments via a simple method of trial and error. The following hyperparameters were used in both cases: a learning rate $\alpha$ of 0.01, a discount factor $\gamma$ of 0.95, an exploration rate $\epsilon$ of 0.99 and a decay rate of 0.995. Within the code provided, the implementation of Q and Double Q-Learning is as described in Fig.~\ref{RL}.  Tables \ref{tabel_results_1} and \ref{tabel_results_2} showcase the consistent supremacy of both Q-Learning and Double Q-Learning for problems ranging from TSP20 to TSP200, with an average difference of approximately 8.2\% to FS and LS. However, performance deteriorates as the number of sampling points escalates, with a noticeable performance gap for TSP500 at 1.36\% for QL and 3\% for Double QL. Evidently, TSP1000 encounters difficulty in escaping local optima, standing as the least efficient of the dataset, diverging more than 6\% from the optimal solution. Furthermore, Table~\ref{tabel_results_std} illustrates that the standard deviation for each TSP instance run is relatively low, ranging from dozens to hundreds of meters. This is minor when compared to the total distance, which extends to kilometers covered by ROMIE. These results indicate that the tours predicted by the RL algorithm are quite consistent across different runs for both RL techniques. The low variability in the results implies that the algorithm performs consistently, yielding similar tour lengths across multiple runs. This consistency is vital for practical applications, where predictability and reliability are crucial. When comparing this algorithm to others, the low standard deviation suggests that any observed differences in performance are likely due to the intrinsic characteristics of the algorithms rather than random variability. This means that the algorithm's performance is stable and dependable, making it a reliable choice for solving TSP instances. Despite various attempts at hyperparameter tuning, this deviation can be attributed to the high computational complexity and the extensive Q-table, which results in incomplete learning and subsequently, subpar solutions. This could also be due to the balance between exploration and exploitation, which might have been skewed towards lower exploration. This is experienced by the rapid epsilon decay applied in the model. The swift convergence is due to a limited number of episodes.  The problem is that Q-Learning is a tabular method meaning that the states and actions spaces are not small enough to be represented efficiently by arrays and tables and therefore it is not scalable for high instances of TSP. \break

\begin{figure}[H]
    \centering
            \subfloat[\scriptsize{Example Of Training Graph For TSP100 Using Q-Learning}]{
          \includegraphics[width=0.75\textwidth]{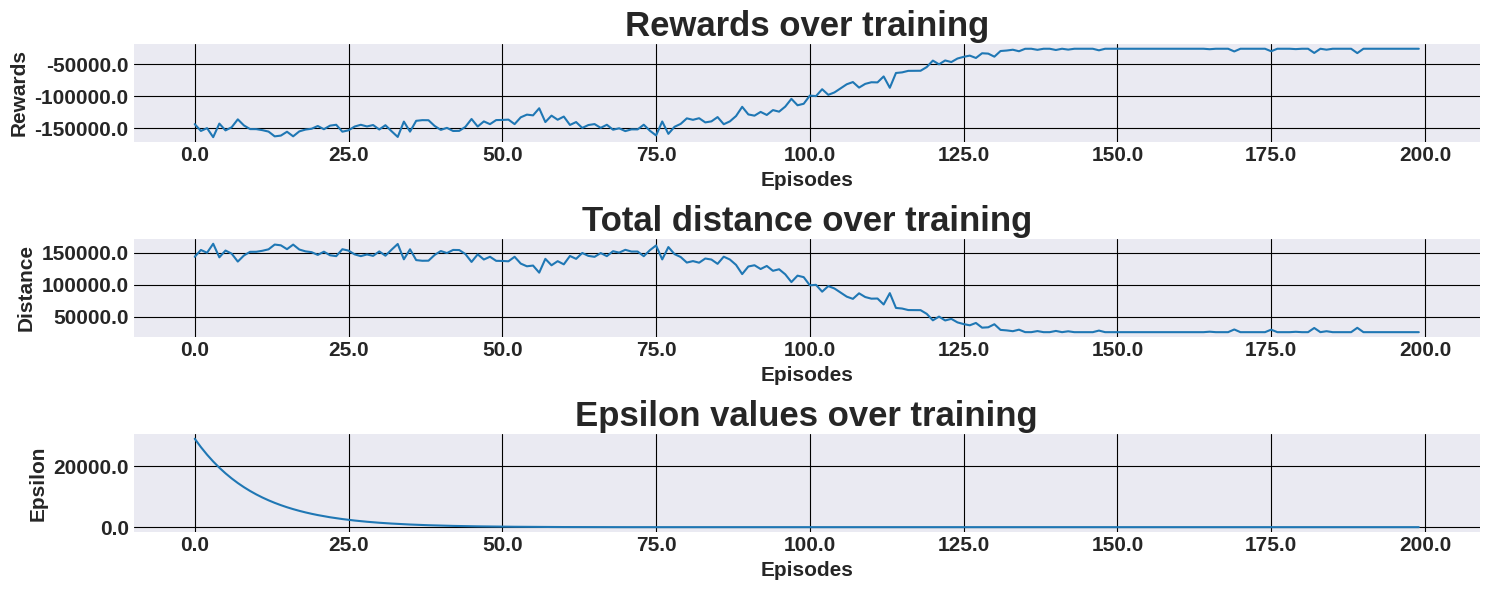}
          \label{fig:QL_50}
          }
        \qquad
        \subfloat[\scriptsize{Example Of Training Graph For TSP1000 Using Q-Learning}]{
          \includegraphics[width=0.75\textwidth]{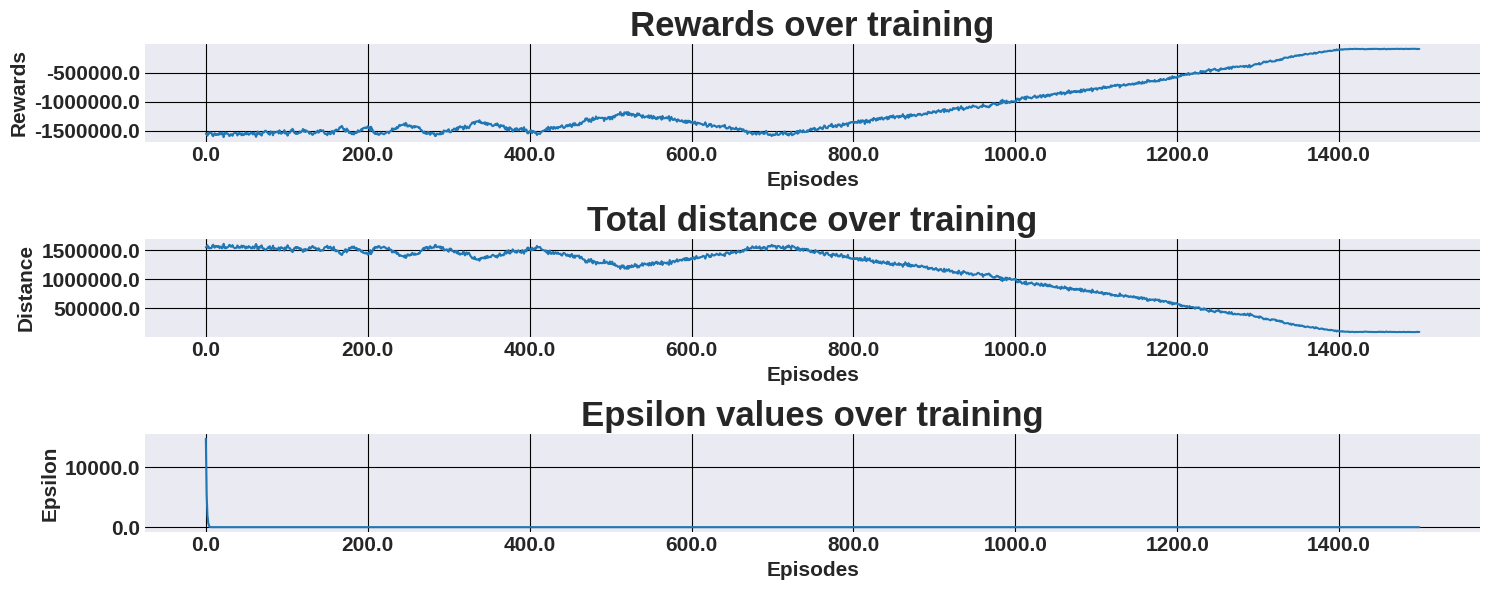}
          \label{fig:QLresultsTSP500}
        }
    \caption{Results For Q-Learning and Double Q-Learning Implementation.}
    \label{fig:QLresults}
\end{figure}

The graphical representation of the iterations against the reward, distance (negative reward), and the decay in epsilon value for the TSP500 dataset can be seen in Fig.~\ref{fig:QLresults}. All training graphs across the dataset exhibit a similar pattern. The triangular pattern observed can primarily be attributed to the alternation between exploration and exploitation phases in the Q-learning algorithm. The algorithm maximizes the current best action during the exploitation phase, leading to an increase in reward. Subsequently, during the exploration phase, it opts for a random action to accumulate new knowledge. However, as this action might not be the most optimal one, the reward may decrease, causing a downward slope in the graph and forming the characteristic triangular pattern. \break

The learning rate, denoted as alpha, plays a pivotal role. This parameter dictates how much the new information will replace the old. A non-optimal learning rate could result in the algorithm 'forgetting' beneficial strategies, causing the reward to drop after reaching a peak. This phenomenon contributes to the triangular shape of the graph. Therefore, to address this, one may need to fine-tune the hyperparameters or provide a stable environment during the training phase to ensure a consistent increase in reward over iterations. \break

\begin{figure}[H]
    \centering
        \subfloat[TSP graph for TSP 20]{
          \includegraphics[width=0.4\textwidth]{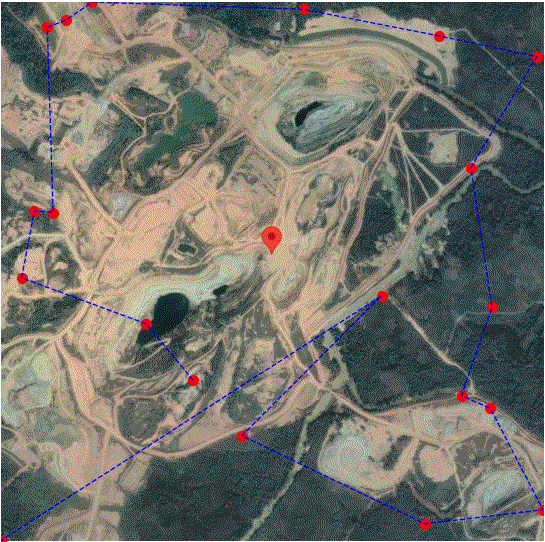}
          \label{fig:tsproute20}
        }
        \qquad
        \subfloat[TSP graph for TSP 50]{
          \includegraphics[width=0.4\textwidth]{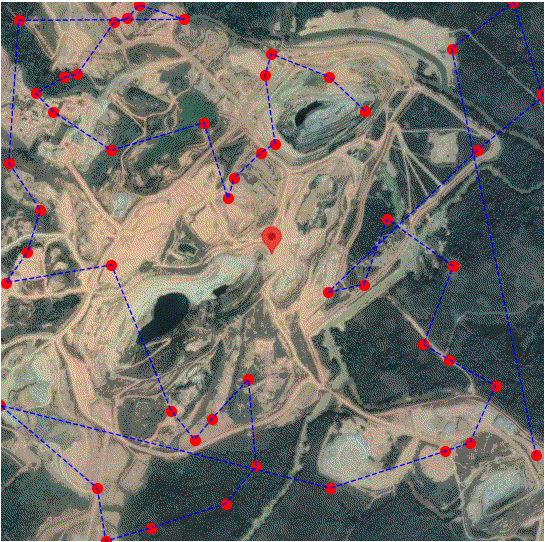}
          \label{fig:tsproute50}
          }
                  \qquad
        \subfloat[TSP graph for TSP 100]{
          \includegraphics[width=0.4\textwidth]{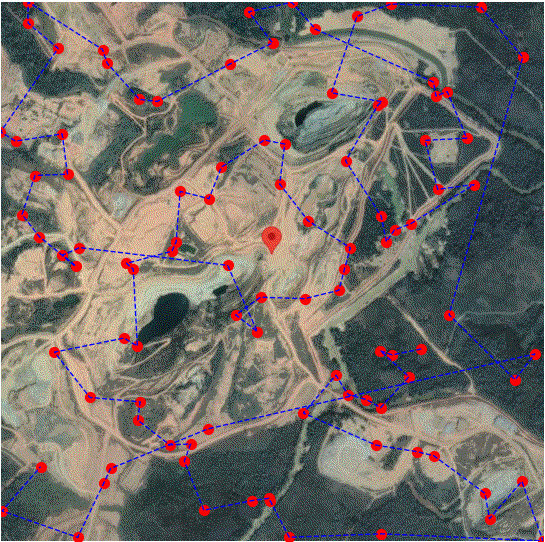}
          \label{fig:tsproute100}
          }
                  \qquad
        \subfloat[TSP graph for TSP 200]{
          \includegraphics[width=0.4\textwidth]{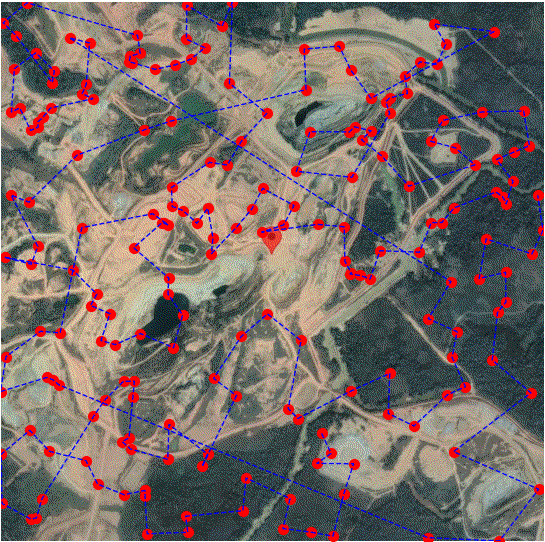}
          \label{fig:tsproute200}
          }
                  \qquad
        \subfloat[TSP graph for TSP 500]{
          \includegraphics[width=0.4\textwidth]{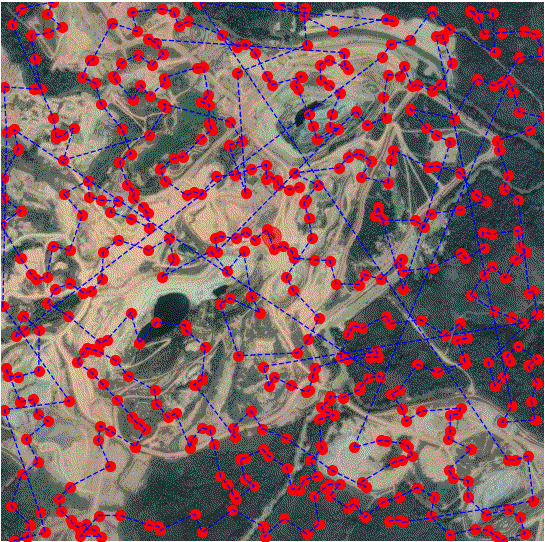}
          \label{fig:tsproute500}
          }
                  \qquad
        \subfloat[TSP graph for TSP 1000]{
          \includegraphics[width=0.4\textwidth]{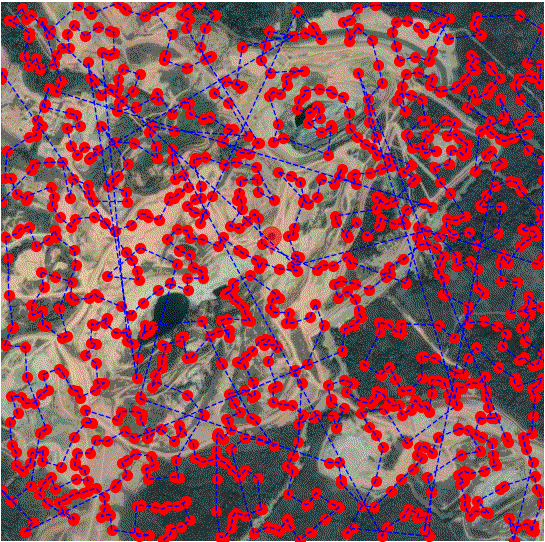}
          \label{fig:tsproute1000}
          }
    \caption{Results of ROMIE way points route generated by Q-learning.}
    \label{fig:TSPQL}
\end{figure}

Fig.~\ref{fig:TSPQL} depicts the physical route that ROMIE will follow for different datasets, as generated by the Q-Learning algorithm. The developed algorithms for the software are available at \href{https://github.com/amgb20/Travelling_Salesman_Problem.git}{Github 1}, \href{https://github.com/amgb20/ROMIE---Global-Path-Planning}{Github 2} and \href{https://github.com/amgb20/Q-Learning-for-TSP/tree/main}{Github 3}.

\subsection{ROMIE GPP Application of The TSP}

This study was carried out in conjunction with the development of the developed GPP software, which resulted in a user-friendly web interface. The interface, built on Django, offers efficient communication between JavaScript and Python files, utilizing the Fetch API for HTTP requests and integrating a security feature called CSRF token. \break

The primary objective was to create an interface that is intuitive for mining engineers, geologists, and researchers. This led to the incorporation of an interactive Google Map. Users can interact with this map in a number of ways, including selecting the mining site's location, delineating the site by dragging on the map (which simultaneously generates the site's coordinates), and choosing sampling points.\break

The users have two options for selecting sampling points: they can either use a grid-based method, or manually click on the map to specify the points where the robot needs to conduct prospecting. Upon the selection of sampling points, the system generates the most efficient TSP route and displays the path that ROMIE will take.\break

\begin{figure}[H]
    \centering
    \includegraphics[width=0.9\textwidth]{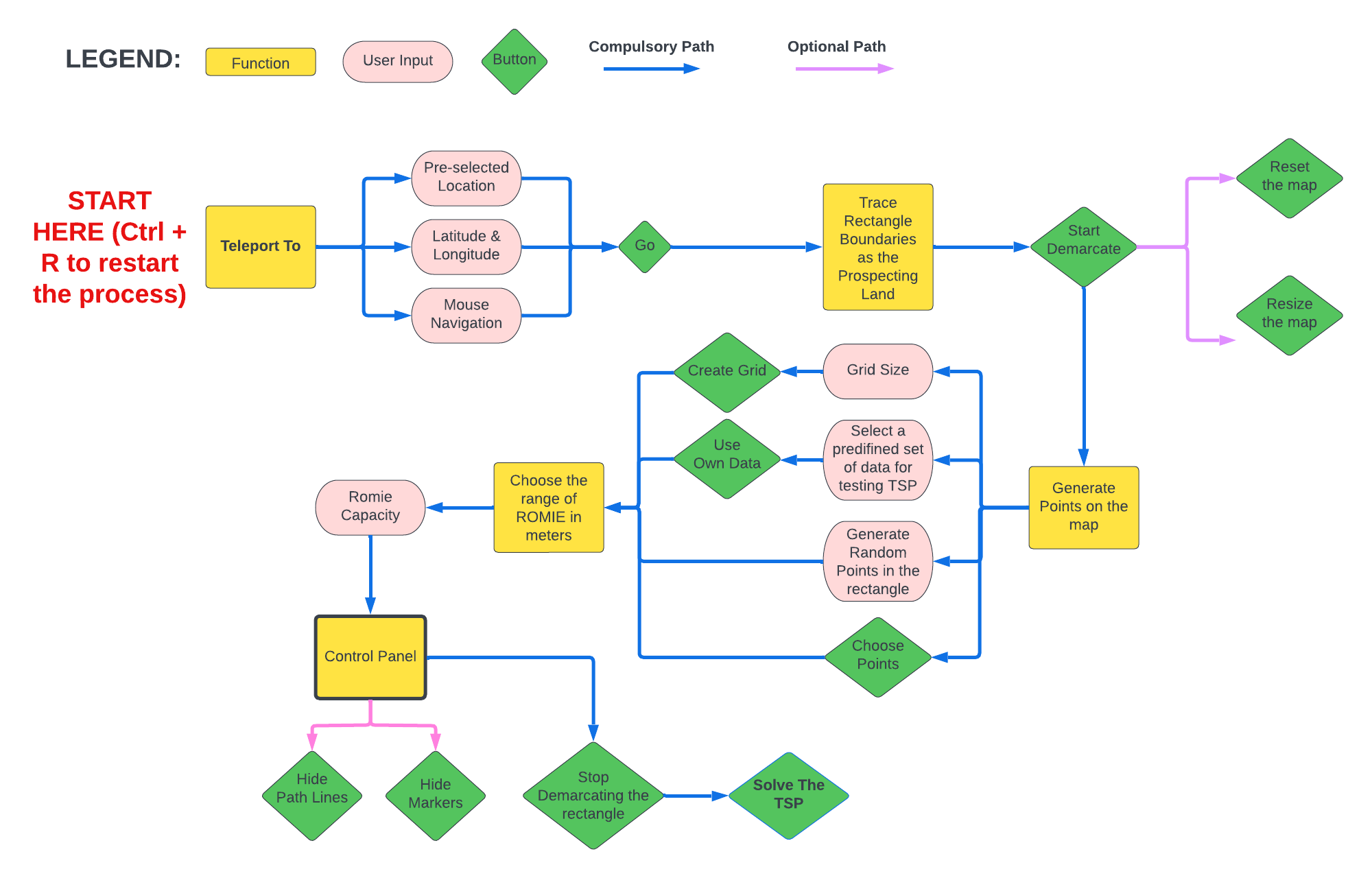}
    \caption{Flow Diagram for User Experience.}
    \label{flowdiagram}
\end{figure}

The interface is organized to have the interactive map on one side, and the user input fields on the other. This layout enhances the user experience by providing a clear and convenient workspace.  The following flow diagram in Fig.~\ref{flowdiagram} indicates the user how to navigate through the web-interface.\break

In essence, this study demonstrates a practical application of solving the TSP problem in the mining industry, bridging the gap between computational solutions and real-world challenges.\break

\begin{figure}[H]
    \centering
    \includegraphics[width=0.9\textwidth]{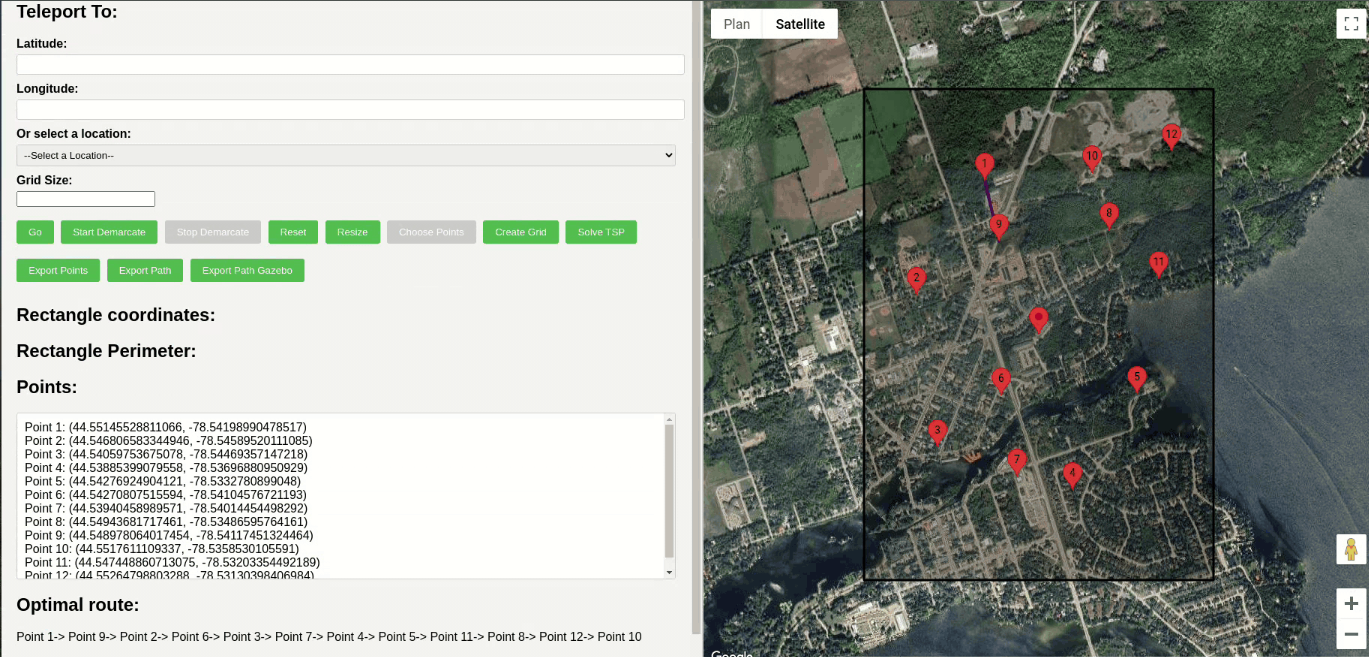}
    \caption{Interface of the GPP software.}
    \label{fig:GPPwebsite}
\end{figure}

\section{Conclusion}

In this study, a new UGV was developed using innovative algorithms such as Q-learning-based Google methods, software practice, and hardware tools. Additionally, we conducted a comparative analysis to evaluate the performance of Google OR-Tools algorithms against RL algorithms, such as Q-Learning and Double Q-Learning. The primary goal of this comparison was to determine if the RL algorithms could surpass the effectiveness of Google OR-Tools in solving the problem. This work offers a comprehensive analysis of the GPP subsystem for UGVs employed as autonomous mining sampling vehicles. The simulation tests were conducted on a data set ranging from 20 to 1,000 sampling points. The achieved results were experimentally validated in this report.

The findings indicate that, on average, the Local Search algorithm in Google OR-Tools deviates from the optimal solution by 5.7\%, compared to the First Solution strategy's deviation of 6.3\%. For sampling points less than 500, Q-Learning exhibits superior performance compared to Double Q-Learning and Google OR-Tools. However, for problems with 500 sampling points, the Guided Local Search algorithm shows superior performance, and for 1,000 points, the Generic Tabu Search algorithm is more effective.

The researchers see a need for future research to explore the application of Deep Neural Networks. Investigating the current state of the art, such as UTSP, or implementing Deep Q-Learning algorithms could potentially yield even better results and seems to be the logical next step in advancing in the realm of ROMIE's GPP software.

\medskip

\tableofcontents

\end{document}